%% file: imr.tex
\documentclass{article}

\usepackage[preprint,nonatbib]{neurips_2020}

\usepackage[utf8]{inputenc} 
\usepackage[T1]{fontenc}    
\usepackage[pagebackref=true,breaklinks=true,letterpaper=true,colorlinks,bookmarks=false]{hyperref}
\usepackage{url}            
\usepackage{booktabs}       
\usepackage{amsfonts}       
\usepackage{nicefrac}       
\usepackage{microtype}      
\usepackage{booktabs}
\usepackage[font=small,labelfont=bf]{caption}
\usepackage{xcolor}
\usepackage{graphicx}
\usepackage{amsmath}
\usepackage{amssymb}
\usepackage{epsfig}
\usepackage{graphicx}
\usepackage{multirow}
\usepackage[numbers,sort]{natbib}
\usepackage{lib}
\usepackage[lofdepth,lotdepth]{subfig}
\usepackage{wrapfig}
\newcommand{\addpic}[1]{\includegraphics[width=8em]{#1}}
\newcommand{\addpicsup}[1]{\includegraphics[height=7em]{#1}}
\title{Implicit Mesh Reconstruction \\ from Unannotated Image Collections}

\author{%
   \qquad Shubham Tulsiani$^1$ \qquad Nilesh Kulkarni$^2$ \qquad Abhinav Gupta$^{1,3}$ \\
$^1$Facebook AI Research \qquad $\,^2$University of Michigan \qquad$\,^3$Carnegie Mellon University \\
{\tt \small \href{https://shubhtuls.github.io/imr/}{https://shubhtuls.github.io/imr/}}
}

\begin{document}
{%
\vspace{-5em}
\maketitle
\vspace{-2em}
\begin{center}
   \centering \includegraphics[width=\textwidth]{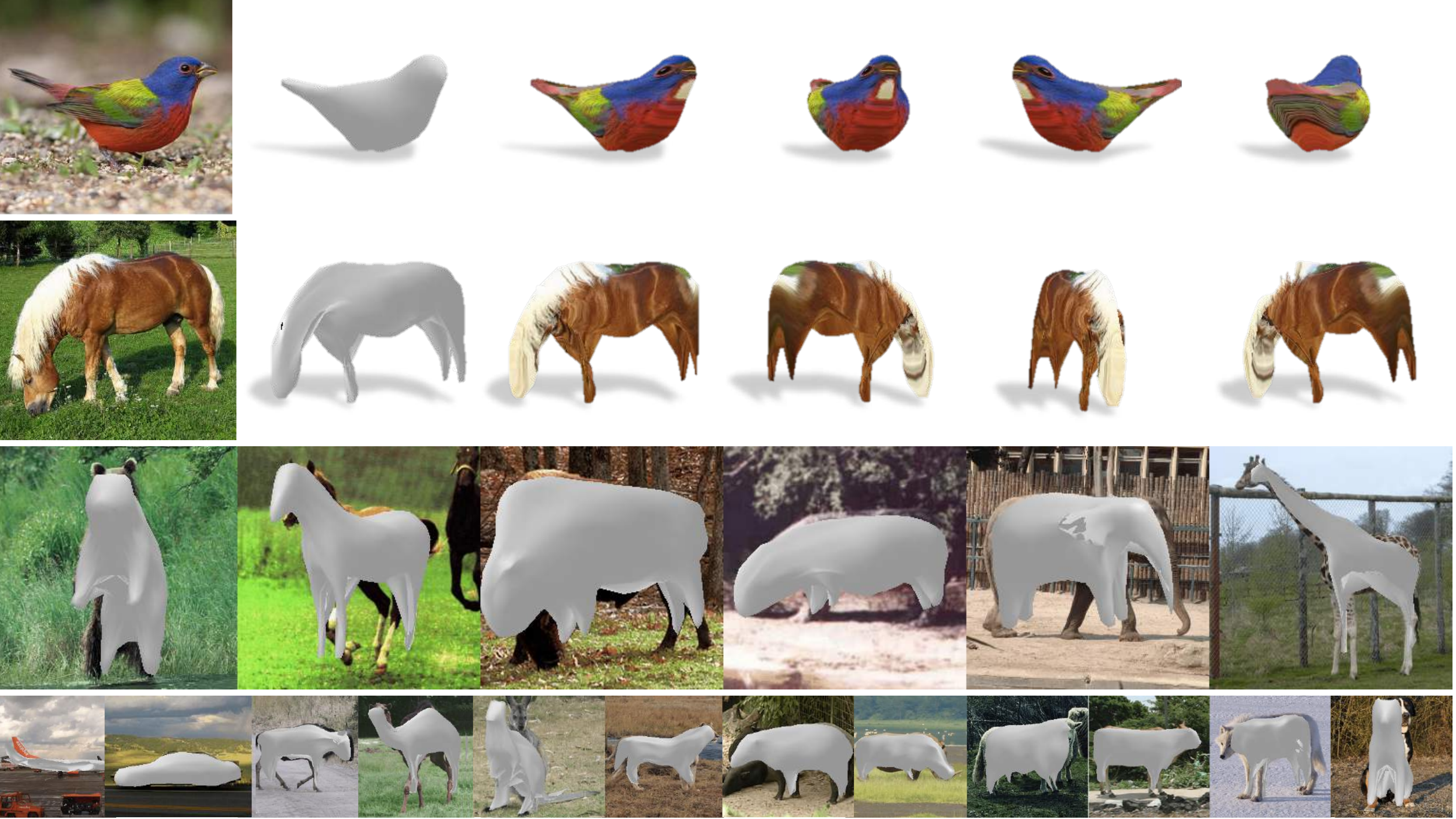} \captionof{figure}{Given a single input image, we can infer the shape, texture and camera viewpoint for the underlying object. In rows 1 and 2, we show the input image, inferred 3D shape and texture from the predicted viewpoint, and three novel viewpoints. We can learn 3D inference using only in-the-wild image collections with approximate instance segmentations, our approach can be easily applied across a diverse set of categories. Rows 3 and 4 show sample predictions across a broad set of categories, with the predicted 3D shape overlaid on the input image. Please see the \href{https://shubhtuls.github.io/imr/}{project page} for 360 degree visualizations.}
   \figlabel{teaser}
\end{center}%
}

\begin{abstract}
We present an approach to infer the 3D shape, texture, and camera pose for an object from a single RGB image, using only category-level image collections with foreground masks as supervision. We represent the shape as an image-conditioned implicit function that transforms the surface of a sphere to that of the predicted mesh, while additionally predicting the corresponding texture. To derive supervisory signal for learning, we enforce that: a) our predictions when rendered should explain the available image evidence, and b) the inferred 3D structure should be geometrically consistent with learned pixel to surface mappings. We empirically show that our approach improves over prior work that leverages similar supervision, and in fact performs competitively to methods that use stronger supervision. Finally, as our method enables learning with limited supervision, we qualitatively demonstrate its applicability over a set of about 30 object categories.
\end{abstract}

\newcommand{\sps}{\mathbb{S}^2}

\input{intro}

\input{related}

\input{approach}

\input{experiments}

\input{discussion}


{\small
\bibliographystyle{ieee}
\bibliography{imr}
}

\clearpage
\input{appendix}

\input{figures_supp}


\end{document}

%% file: intro.tex
\section{Introduction}
\vspace{-2mm}
Inferring 3D structure of {\bf diverse} categories of objects from {\bf images in the wild} (see \figref{teaser}) has been one of the long term goals in computer vision. Despite the decades of progress in computing, graphics and machine learning, we do not yet have systems that can infer the underlying 3D for objects in natural images. This is in stark contrast to the progress witnessed in related problems such as object recognition and detection where we have developed scalable methods that make accurate predictions for hundreds of categories for images in the wild. Why is there this disconnect between progress in 3D perception and 2D understanding, and what can we do to overcome it? We argue that the central bottleneck for 3D perception in the wild has been the strong reliance on 3D supervision, or the low expressivity of previous models (e.g., fixed 3D template). In this work, we aim to bypass these bottlenecks and present an approach that can learn inference of a deformable 3D shape using only the form of data that 2D recognition systems leverage -- in the wild category-level image collections with (approximate) instance segmentations. 

Given a single input image, our goal is to be able to infer the 3D shape, texture and camera pose for the underlying object. A scalable solution needs to have two ingredients: (a) 3D modeling which can handle instance variations and deformations of objects; (b) require minimal supervision to allow scalability across diverse categories. There have been recent attempts on both these axes. For example, a recent approach~\cite{cmrKanazawa18} learns explicit 3D representations from image collections, and can handle instance variations and pose variations. But this approach crucially relies on 2D keypoint annotations to guide learning, thereby making it difficult to scale beyond a handful of categories. On the other hand, Kulkarni  \etal~\cite{kulkarni2019canonical,kulkarni2020articulation} learn pixel to surface mappings that are consistent with a global (articulated) 3D template, and show that this can help learn accurate prediction. This approach bypasses keypoint supervision, but cannot model any shape  variations (fat vs thin bird) and additionally requires 3D part supervision for modeling articulations.

Our approach handles both, shape and pose variation by inferring implicit category-specific shape representations. To bypass the need for direct supervision, we also infer the corresponding texture and enforce reprojection consistency between our predictions and the available images and segmentations. Drawing inspiration from the work by Kulkarni \etal~ \cite{kulkarni2019canonical}, we use a geometric cycle-consistency loss between global 3D and local pixel to surface predictions to derive additional learning signal from unannotated image collections. Leveraging a single 3D template shape per category as initialization, we learn the category-level implicit shape space from image collections. We show sample shape predictions obtained using our approach in \figref{teaser} and also visualize the 3D shape with predicted texture from the predicted and novel camera viewpoints. We observe that, despite the lack of direct supervision, our method effectively captures the shape variation across instances \eg head bending down, length of tail. As illustrated in \figref{teaser}, our approach is applicable across a diverse set of object categories, and the reliance on only image collections with approximate segmentation masks allows learning in settings where previous approaches could not.

%% file: related.tex
\vspace{-2mm}
\section{Related Work}
\vspace{-2mm}

\paragraph{Learning 3D Shape from Annotated Image Collections.} 
The recent success of deep learning has resulted in a number of learning based approaches for the task of single-view 3D inference. The initial approaches ~\cite{choy20163d,girdhar2016learning} showed impressive volumetric inference results using synthetic data as supervision, and these were then generalized to other representations such as point clouds~\cite{fan2017point}, octrees~\cite{tatarchenko2017octree, hane2017hierarchical}, or meshes~\cite{wang2018pixel2mesh}.
However, these approaches relied on ground-truth 3D as supervision, and this is difficult to obtain at scale or for images in the wild. There have therefore been attempts to relax the supervision required \eg by instead using multi-view image collections~\cite{rezende2016unsupervised,yan2016perspective,drcTulsiani17,mvcTulsiani18,tatarchenko2016multi}. Closer to our setup, several approaches~\cite{cmrKanazawa18,kar2015category, cashman2013shape,chen2019learning} have also addressed the task of learning 3D inference using only single-view image collections, although relying on additional pose or keypoint annotations.
While these approaches yield encouraging results, their reliance on annotated 2D keypoints limits their applicability for generic categories.
Our work also similarly learns from image collections, but does so without using any supervision in the form of 2D keypoints or pose; and this allows  us to learn from image collections of objects in the wild.

\vspace{-2mm}
\paragraph{Learning 3D from Unannotated Image Collections.}
Our motivation of learning 3D from unannotated image collections is also shared by some recent works~\cite{kulkarni2019canonical, kulkarni2020articulation, nguyen2019hologan}. Nguyen-Phuoc \etal ~\cite{nguyen2019hologan} use geometry-driven generative modeling to learn 3D structure, but their approach only infers a volumetric feature capable of view synthesis. Unlike our work, this does not output a tangible 3D shape or texture for the underlying object. Closer to our approach, Kulkarni \etal~\cite{kulkarni2019canonical, kulkarni2020articulation} predict explicit 3D representations by proposing a cycle-consistency loss between inferred 3D and learned pixelwise 2D to 3D mappings. However, their method only produces a limited 3D representation in the form of a rigid or articulated template, and cannot handle intra-instance shape variations \eg fat vs thin bird. Our implicit shape representation allows capturing such variation in addition to the instance texture, and we generalize their cycle consistency loss for our representation.

\vspace{-2mm}
\paragraph{Implicit 3D Representations.} 
Several recent works~\cite{mescheder2019occupancy, park2019deepsdf, xu2019disn, saito2019pifu} using implicit functions have shown impressive results on the tasks of 3D reconstruction. Unlike explicit representations (e.g. meshes, voxels,  point clouds) these methods learn functions to parameterize a 3D volume or surface.
Our representation is inspired by work from Groueix \etal ~\cite{groueix2018papier} which learns a mapping conditioned on the latent code from points on a 2D manifold to a 3D surface, and we further equip this with an implicit texture as in Oechsle~\etal ~\cite{oechsle2019texture}. However, all these prior approaches require settings with ground-truth 3D and texture supervision available. In contrast, our work leverages these representations in an unsupervised setup with corresponding technical insights to make learning feasible \eg use of a `texture flow' and incorporation of category-specific shape consistency.

%% file: approach.tex
\vspace{-2mm}
\section{Approach}
\vspace{-2mm}
Given an image of an object, we aim to infer its 3D shape, texture, and camera pose. Moreover, we want to learn this inference using only image collections with instance segmentations as supervisory signal. Towards this, our approach leverages implicit shape and texture representations, and enforces geometric consistency with available image to bypass the need of direct supervision. 

Specifically, we leverage geometry-driven objectives to learn an encoder which predicts the desired properties from the input image $f_{\theta'}(I) \equiv (\pi, \mathbf{z_s}, \mathbf{z_t})$ -- here $\pi$ denotes a weak perspective camera, and $\mathbf{z_s}, \mathbf{z_t}$ correspond to latent variables that instantiate the underlying shape and texture. We first describe in \secref{representation} the category-specific implicit shape representation pursued and then describe the proposed learning framework in \secref{learning}. While we initially consider only shape inference, we show in \secref{texture} how our approach can also incorporate texture prediction.


\vspace{-2mm}
\subsection{Category-specific Implicit Mesh Representation}
\vspace{-2mm}
\begin{figure}[h!]
    \centering
    \begin{minipage}{0.45\textwidth}
        \centering
        \includegraphics[width=0.9\textwidth]{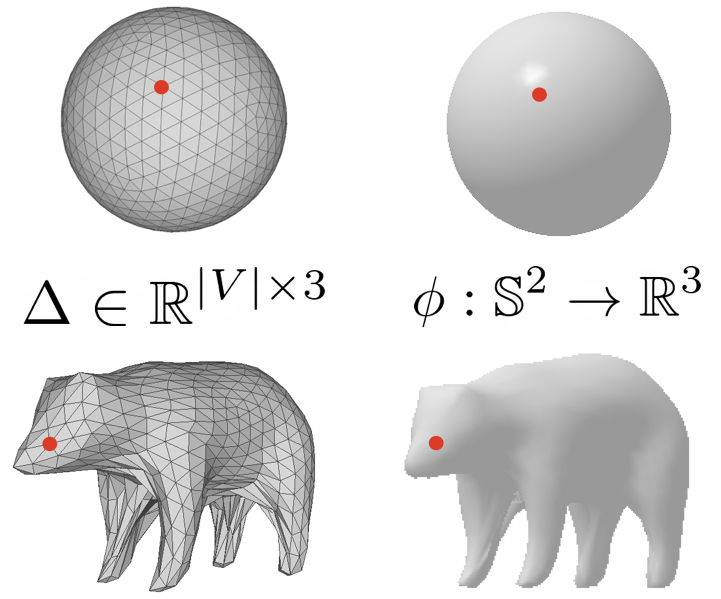} 
        \caption{\textbf{Mesh Parametrization.} An explicit representation (left) of deformation of a sphere to a mesh is parametrized via the location of a fixed set of 3D vertices. In contrast, an implicit representation (right) is parametrized via a function that maps any point on a sphere to a 3D coordinate.}
        \figlabel{expimp}
    \end{minipage}\hfill
    \begin{minipage}{0.52\textwidth}
        \centering
        \includegraphics[width=0.98\textwidth]{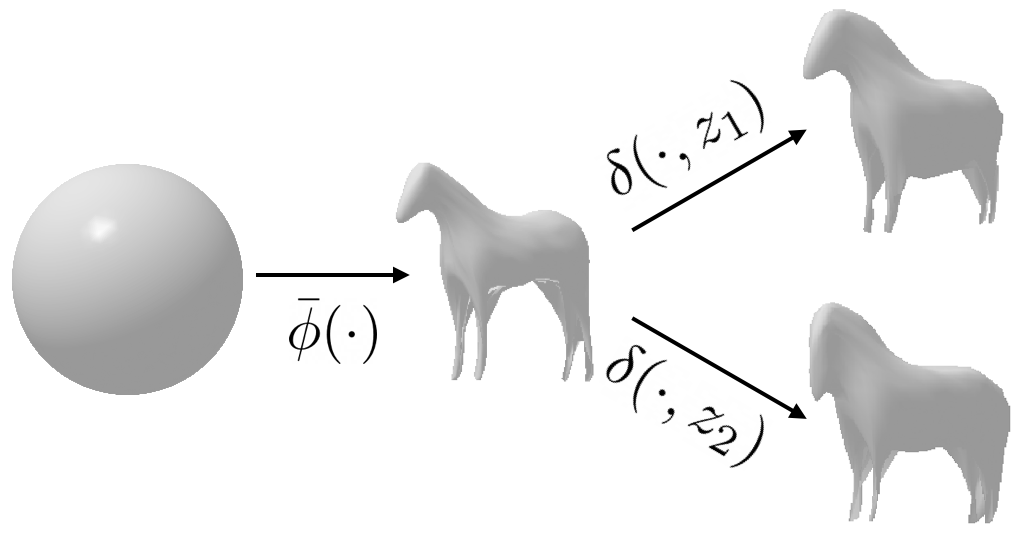} 
        \caption{\textbf{Category-specific Implicit Mesh Representation.} We represent the shapes for different instances in a category via a latent-variable conditioned implicit function. The mesh for a given instance is obtained by adding an instance-specific deformation to a shared category-level mean shape, where both, the shared template and the deformations, are represented as neural networks.}
        \figlabel{csimp}
    \end{minipage}
\end{figure}

\seclabel{representation}
We model a predicted shape as a deformation of a unit sphere. One possible representation for such a deformation is an \emph{explicit} one: we can represent the unit sphere as a mesh with $V$ vertices, and parametrize its deformation as a per-vertex translation $\delta \in \mathbb{R}^{|V| \times 3}$. While several prior works~\cite{cmrKanazawa18,kato2018neural} have successfully leveraged this explicit representation, it is computationally challenging to scale to a finer mesh and also lacks certain desirable inductive biases \eg correlation of vertex locations, as these are parametrized independently. To overcome these challenges, we instead \emph{implicitly} parametrize the shape (see \figref{expimp}). Denoting by $\sps$ the surface of a unit sphere, we parametrize its deformation via a function $\phi: \sps \rightarrow \mathbb{R}^3$, such that for all $\mathbf{u} \in \sps$, $\phi(\mathbf{u})$ is a point on the implied surface.

However, a given $\phi$ as defined above only represents a specific shape, whereas we are interested in modeling the different possible shapes across instances of a category. We therefore model the implicit shape function as additionally being dependent on a latent variable $\mathbf{z_s} \in \mathbb{R}^d$, such that $\phi(\cdot, \mathbf{z_s}): \sps \rightarrow \mathbb{R}^3$ can describe different shapes according to the variable $\mathbf{z_s}$. While the instance-specific latent variable allows representing the variation within the category, we factorize $\phi$ to also benefit from the commonalities within it -- we model the instance level shape as a combination of a instance-agnostic mean shape and an instance dependent deformation (see \figref{csimp}).
\begin{gather}
    \phi(\mathbf{u}, \mathbf{z_s}) = \bar{\phi}(\mathbf{u}) + \delta(\mathbf{u}, \mathbf{z_s}); ~~ \mathbf{u} \in \sps; \mathbf{z_s} \in \mathbb{R}^d
\end{gather}
Here $\bar{\phi}(\cdot)$ and $\delta(\cdot, \cdot)$ are both modeled as neural networks, and represent the implicit functions for a category-level mean shape and instance level deformation respectively. To overcome ambiguities and possible degenerate solutions when learning without supervision, we initialize $\bar{\phi}(\cdot)$ for each category to match a manually chosen template 3D shape (see appendix for details and visualizations).



\vspace{-2mm}
\paragraph{Enforcing Symmetry.} Almost all naturally occurring objects, as well as several man-made ones, exhibit reflectional symmetry. We incorporate this by constraining the learnt mean shape $\bar{\phi}$ and deformations $\delta(\cdot, \mathbf{z_s})$ to be symmetric along the $X=0$ plane.
For example, the deformation is constrained to be symmetric as follows (where $\mathcal{R}(\cdot)$ denotes a reflection function):
\begin{gather}
    \delta(\mathbf{u}, \mathbf{z_s}) \equiv (~\delta'(\mathbf{u}, \mathbf{z_s}) + \mathcal{R}(\delta'(\mathcal{R}(\mathbf{u}), \mathbf{z_s}))~)/2;
\end{gather}

\vspace{-2mm}
\paragraph{Encouraging Locally Rigid Transforms.}
In our formulation, the instance-specific deformation $\delta(\cdot, \mathbf{z_s})$ is required to capture any change from the category-level mean shape. This change in 3D structure can stem from intrinsic variation \eg a bird can be thin or fat, or can be caused by articulation \eg the same horse would induce different deformations if its head is bent down or held upright. As articulations can be viewed as local rigid transformations of the underlying shape, we encourage the learnt deformations to explain the variation using locally rigid transformations if possible. We note that under any rigid transform, the distance between two corresponding points remains unchanged, and incorporate a regularization that penalizes the mean of this variation across local neighborhoods. Denoting by $\mathcal{N}(\mathbf{u})$ a local neighborhood of $\mathbf{u}$, this objective is:
\begin{gather}
L_{\text{rigid}} =  \underset{\mathbf{u}  \in \mathbb{S}^2}{\mathop{{}\mathbb{E}}} ~~
\underset{\mathbf{u'}  \in \mathcal{N}(\mathbf{u})}{\mathop{{}\mathbb{E}}} |~ \| \phi(\mathbf{u},\mathbf{z_s}) - \phi(\mathbf{u'},\mathbf{z_s})  \| - \| \bar{\phi}(\mathbf{u}) - \bar{\phi}(\mathbf{u'})  \| ~|
\end{gather}

\vspace{-2mm}
\subsection{Learning 3D Inference via Reprojection Consistency}
\vspace{-2mm}
\seclabel{learning}
While we do not have direct supervision available for learning the shape and pose inference, we can nevertheless derive supervisory signal by encouraging our 3D predictions be geometrically consistent with the available image evidence. Concretely, we enforce that the predicted 3D shape, when rendered according to the predicted camera, matches the foreground mask, while specially emphasizing boundary alignment. Following a surface mapping consistency formulation by Kulkarni \etal~\cite{kulkarni2019canonical}, we also implicitly encourage semantically similar regions across instances to be `explained' by consistent regions of the deformable shape space.

\vspace{-2mm}
\paragraph{Mask and Boundary Reprojection Consistency.} Given the inferred (implicit) 3D shape $\phi_{\theta}(\cdot, \mathbf{z_s})$, we recover an explicit mesh $M$ by sampling the implicit function at a fixed resolution. We then use a differentiable renderer~\cite{kato2018neural, ravi2020pytorch3d} to obtain the foreground mask for this predicted mesh   camera $\pi$, and define a loss against the ground-truth mask $I_f$: $L_{\text{mask}} = \|I_f - f_{render}(M, \pi)\|_1$.

While this loss coarsely aligns the predictions to the image, it does not emphasize details such as long tails of birds, or animal legs. We therefore  incorporate an objective  $L_{\text{boundary}}$ as proposed by Kar \etal~\cite{kar2015category} --  points should project within the foreground, and  contour points should have some projected 3D points nearby (see appendix for formulation). 


\vspace{-2mm}
\paragraph{Regularization via Pixel to Surface Mappings.} In our formulation, each point on the predicted 3D shape corresponds to a unique point $\mathbf{u}$ on the surface of a unit sphere. To ensure that the inferred 3D is consistent across different instances in a category, we would ideally like to enforce that semantically similar regions across images are `explained' by similar regions of the unit sphere \eg the $\mathbf{u}$ projecting on the horse head is similar across instances. However, we do not have access to any semantic supervision to directly operationalize this insight, but we can do so indirectly.

Following the work of Kulkarni \etal~\cite{kulkarni2019canonical}, we train a convolutional predictor to infer pixel to surface mappings $C \equiv g_{\Theta}(I)$ given an input image $I$. The inferred mapping $C[\mathbf{p}] \in \mathbb{S}^2$ predicts for each pixel $\mathbf{p}$ a corresponding coordinate on the unit sphere, and thereby the 3D shape, and we encourage this prediction to be cycle-consistent with the inferred 3D. Intuitively, the convolutional predictor, using local image appearance, predicts what 3D point each pixel corresponds to. Therefore enforcing consistency between the inferred 3D and the predicted mapping implicitly encourages our inferred 3D to be semantically consistent. Concretely, via the predicted mapping, a pixel $\mathbf{p}$ is mapped $C[\mathbf{p}] \in \sps$, and thus to the 3D point $\phi(C[\mathbf{p}], \mathbf{z_s})$ on the implicit mesh, and we encourage its reprojection under camera $\pi$ to map back to the pixel.  While~\cite{kulkarni2019canonical} learned these mappings in context of a rigid or rigged template, we can extend their loss formulation for our implicit mesh representation.
\begin{gather}
   L_{\text{gcc}} = \underset{\mathbf{p}}{\sum} \| \pi(\phi(C[\mathbf{p}],\mathbf{z_s})) - \mathbf{p}\|
\end{gather}

\vspace{-4mm}
\paragraph{Leveraging Optional Keypoint Supervision.} While our goal is to learn 3D reconstruction with minimal supervision, our approach easily allows using semantic keypoint labels \eg nose, tail \etc if available. For each semantic keypoint $k$ annotated in the 2D images, we first manually annotate a corresponding 3D keypoint on the category-level template mesh. This allows us to associate a unique spherical coordinate $\mathbf{u}_k$ with each keypoint $k$ for a category. Given an input training image with annotated 2D keypoint locations $\{x_k\}$, we can penalize the reprojection error.
\begin{gather}
    L_{\text{kp}} = \underset{k}{\sum} \| \pi(\phi(\mathbf{u}_k, \mathbf{z_s})) - x_k \|
\end{gather}
Note that we only leverage this supervision in certain ablations, and that all other results (including visualized predictions) are obtained \textbf{without} using this additional signal \ie only mask supervision.

\vspace{-2mm}
\paragraph{Training Details.} Our reprojection and cycle consistency losses depend on both, the predicted camera $\pi$ and the inferred shape $\phi(\cdot, \mathbf{z_s})$. Unfortunately, learning these together is susceptible to local minima where only a narrow range of poses are predicted. We follow suggestions from prior work ~\cite{mvcTulsiani18,insafutdinov2018unsupervised}  to overcome this and predict multiple diverse pose hypotheses and their likelihoods, and minimize the expected loss. Additionally, as the inferred poses in initial training iterations are often inaccurate, the learned deformations are not meaningful. We therefore only allow training $\delta$ after certain epochs.  We use a pretrained ResNet-18~\cite{he2016deep} based encoder as $f_{\theta'}$, and a simple 4 layer feedforward MLP to instantiate the shape and texture implicit functions.

\begin{figure}[t!]
   \centering \includegraphics[width=1.0\textwidth]{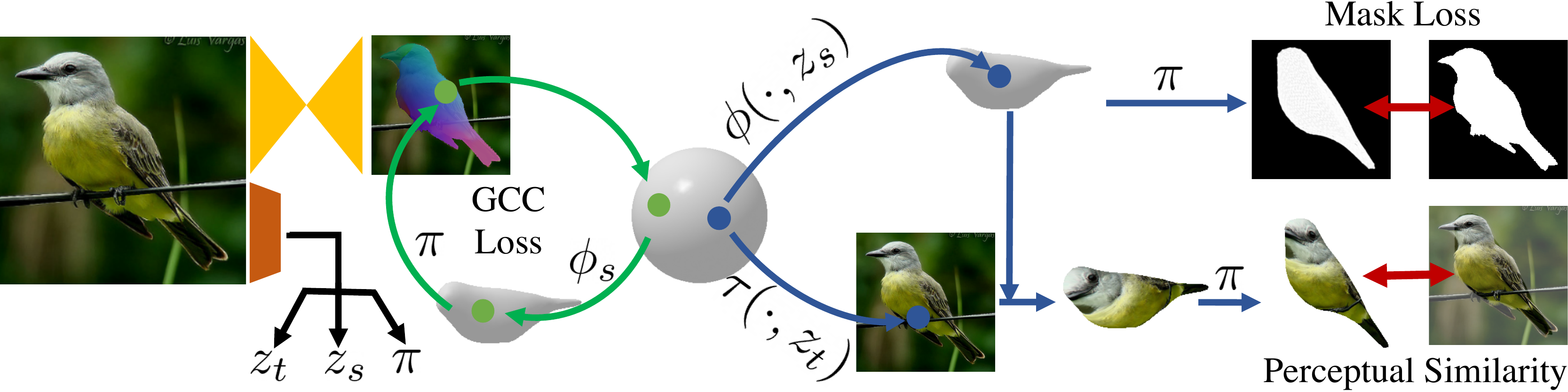}
   \caption{{\bf Overview of Training Procedure.} We train a network to predict the shape, texture, and camera pose for an object in an image, where the shape and texture are parametrized via latent-variable conditioned implicit functions. We learn this network using a combination of reprojection consistency losses between the inferred 3D and the input image and a cycle consistency objective with learned pixel to 3D mappings (see text for details). 
   }
   \figlabel{approach}
\end{figure}

\vspace{-2mm}
\subsection{Texture Prediction via Implicit Texture Flow}
\vspace{-2mm}
\seclabel{texture}
Towards capturing the appearance of the depicted object, we predict the texture that gets overlaid on the predicted (implicit) 3D shape, and learn this inference by enforcing consistency with the observed foreground pixels. As our 3D shape is parametrized via a deformation of a sphere, we simply need to associate each point on the sphere with a corresponding texture to induce a textured 3D shape. 

While related approaches~\cite{cmrKanazawa18} create a fixed resolution texture map, we instead propose to infer an implicit texture function $\tau(\cdot,\mathbf{z_{t}})$ s.t. $\tau(\mathbf{u},\mathbf{z_t})$ yields the texture corresponding to $\mathbf{u}$ given the instance-specific texture encoding $\mathbf{z_t}$. Instead of directly regressing to the color value, we follow Zhou \etal's~\cite{zhou2016view} insight of copying the pixel value, and predict an implicit texture flow: \ie $\tau(\mathbf{u},\mathbf{z_t}) \in \mathbb{R}^2$ indicates the coordinate of the pixel whose color should be copied (via bilinear sampling) to get texture at $\mathbf{u}$. As with our implicit shape representation, we also symmetrize the predicted texture flow function to propagate textures on unseen regions.

To derive learning signal for this prediction, we follow Kanazawa \etal~\cite{cmrKanazawa18} and differentiably render~\cite{liu2019soft} the predicted 3D shape with the implied texture and penalize a perceptual loss~\cite{zhang2018perceptual} against the foreground pixels of the image. We additionally also incorporate a term encouraging the texture flow to sample from foreground pixels instead of background ones.

%% file: experiments.tex
\vspace{-2mm}
\section{Experiments}
\vspace{-2mm}
\subsection{Datasets and Training Setup}
\vspace{-2mm}
\input{tables/kpeval}

We present empirical and qualitative results across a diverse set of categories, and leverage several existing datasets to obtain the required training data. Across all these datasets, we only use the images and annotated (or automatically obtained) segmentation masks for training, but use the additional available annotations \eg keypoints or approximate 3D for evaluation. We download a representative template model (used to initialize $\bar{\phi}$) for all the categories from ~\cite{free3d}. 

\vspace{-2mm}
\paragraph{Rigid Objects} \hspace{-3mm} \emph{(Aeroplane, Car).} We use images from the PASCAL3D+~\cite{xiang2014beyond} dataset, which combines images from PASCAL VOC~\cite{everingham2015pascal} and Imagenet~\cite{russakovsky2015imagenet}. For the former set, a manually annotated foreground mask is available, whereas an automatically obtained one is used for the Imagenet subset. We follow the splits used in prior work~\cite{drcTulsiani17}, but unlike them, do not use the keypoint annotations for training. As annotations for approximate 3D models are available on this dataset, it allows us to empirically measure the reconstruction accuracy.

\vspace{-2mm}
\paragraph{Curated Animate Categories} \hspace{-3mm} \emph{(Bird, Cow, Horse, Sheep).} We use the CUB-200-2011~\cite{WelinderEtal2010} dataset for obtaining bird images with segmentation masks. For the other categories, we use the splits from \cite{kulkarni2019canonical} which, similar to PASCAL3D+, combine images from VOC and Imagenet. Across all these categories, we have keypoint annotations available for images in the test set, and we use these for indirect evaluation of the quality and consistency of the inferred 3D.

\vspace{-2mm}
\paragraph{Quadrupeds from Imagenet} \hspace{-3mm} \emph{(Lion, Bear, Elephant, and 20 others).} We also apply our method on categories from Imagenet using automatically obtained segmentations~\cite{kirillov2019pointrend}. We use the images from Kulkarni \etal~\cite{kulkarni2020articulation} who (noisily) filter out instances with significant truncation and occlusion.

\input{figures.tex}

\vspace{-2mm}
\subsection{Evaluation using Semantic Keypoints}
\vspace{-2mm}
\seclabel{kpeval}
Recall that each point on our predicted 3D mesh corresponds to some $\mathbf{u} \in \sps$. A rendering of this predicted 3D according to the predicted camera therefore induces a per-pixel mapping to $\sps$ (see appendix for visualization). To indirectly evaluate the accuracy and consistency of our inferred 3D, we can use this mapping to infer and evaluate dense 2D to 2D correspondence across images. Following prior work, we report a keypoint-transfer accuracy `PCK-T' (see ~\cite{kulkarni2020articulation} for metric details) in \tableref{kpeval} a). We observe that our approach significantly improves over prior work that in both supervision settings -- with and without keypoint annotations.
Using the annotations of 3D keypoints on the category-level template (see \secref{learning}), we can predict their 2D locations in an image as $\pi(\phi(\mathbf{u}_k, \mathbf{z_{s}})) $. This allows us to compute a reprojection accuracy `PCK-R' (percentage of keypoints reprojected within a certain threshold of ground-truth) for annotated semantic keypoints. We report this accuracy in \tableref{kpeval} b), and observe similar trends. In particular, we note that our method allows more accurate reprojection and transfer than previous approaches that use similar supervision but only infer a restricted 3D representation~\cite{kulkarni2019canonical,kulkarni2020articulation}.

\renewcommand{\arraystretch}{1.2}
\setlength{\tabcolsep}{4pt}
\begin{wraptable}{r}{50mm}
\vspace{-10mm}
\footnotesize
\centering
\begin{tabular}{l c c c}
    \toprule
    Supv & Method & Aero & Car  \\ \midrule
    \multirow{2}{0.7cm}{KP + Mask}
    & CSDM~\cite{kar2015category}  & 0.40 & 0.60 \\
    & DRC~\cite{drcTulsiani17}  & 0.42 & 0.67 \\
    & CMR~\cite{cmrKanazawa18}  & 0.46 & 0.64 \\ 
     \midrule
    Mask & IMR (ours) & 0.44  & 0.66 \\ 
     \bottomrule
\end{tabular}
\caption{\textbf{3D Reconstruction Accuracy}.
We report the mean intersection over union (IoU) of our 3D predictions with the available 3D annotation on PASCAL3D+ dataset. We observe competitive performance with prior approaches that require stronger supervision.}
\tablelabel{ioueval}
\end{wraptable}
\subsection{3D Reconstruction Accuracy}
In \secref{kpeval}, we compared our approach to previous works that learn using similar supervision, but only infer a constrained 3D representation. However, there have also been prior approaches which, similar to ours, allow more expressive 3D inference, but require additional keypoint or pose supervision. We compare our learned 3D inference to these (relatively) strongly supervised methods using the PASCAL3D+ dataset which has approximate 3D ground-truth available in the form manually selected templates.

We report in \tableref{ioueval} the mean intersection over union (IoU) of the predicted 3D shape with the available ground-truth on held out test images. We compare our approach to a deformable model fitting~\cite{kar2015category}, a volumetric prediction~\cite{drcTulsiani17}, and an explicit mesh inference~\cite{cmrKanazawa18} approach -- all of which rely on keypoint or pose labels for learning. We observe that across both the examined categories, our approach, using only foreground mask as supervision, yields competitive (and sometimes better) performance.


\vspace{-2mm}
\subsection{Learning from Unannotated Image Collections}
\vspace{-2mm}
To allow direct or indirect empirical evaluation, we so far only examined categories where annotated image collections are available. However, as we do not leverage these annotations for learning, we can go beyond this limited set of classes and apply our approach to generic categories from Imagenet. In particular, we consider several quadruped categories \eg lion, bear, zebra \etc, and train our approach with automatically obtained approximate instance segmentations using PointRend ~\cite{kirillov2019pointrend}. We visualize our predictions in \figref{qual-results-detection} and also show additional \emph{random} results in the supplementary. We see that our predictions can capture variation \eg thin or fat body, head bending down, sitting vs standing \etc, and that the inferred texture for visible and invisible regions is also meaningful.


\renewcommand{\arraystretch}{1.2}
\setlength{\tabcolsep}{4pt}
\begin{wraptable}{r}{50mm}
\vspace{-1cm}
\footnotesize
\centering
\begin{tabular}{l c c}
    \toprule
    Method & PCK-P & PCK-T \\ \midrule
    Ours  & 54.1 & 41.3 \\
    Ours - $L_\text{gcc}$  & 50.5 & 40.2 \\
    Ours - $L_\text{boundary}$  & 53.6 & 40.3 \\
    Ours - $L_\text{rigid}$  & 51.6 & 41.7 \\  
        \bottomrule
\end{tabular}
\caption{\textbf{Ablation of loss components.} We report mean keypoint reprojection and transfer accuracy across categories when removing certain terms from the objectives. Higher is better.}
\tablelabel{ablations}
\end{wraptable}
\vspace{-2mm}
\subsection{Ablations}
\vspace{-2mm}
We ablate various terms in our learning objective using the semantic keypoint reprojection (PCK-P) and transfer (PCK-T) metrics. In particular, we examine whether incorporating consistency loss with pixelwise surface mappings is helpful, and if prioritizing locally rigid transforms and boundary alignment is beneficial. We report in \tableref{ablations} the mean accuracy for both evaluations across the four animate categories evaluated (bird, horse, cow, and sheep). We observe that all the components improve performance, though we note that some components \eg encouraging rigidity or boundary alignment lead to more significant qualitative improvements.


%% file: tables/kpeval.tex
\renewcommand{\arraystretch}{1.05}
\setlength{\tabcolsep}{3pt}
\begin{table}[t]
\centering
\footnotesize
\centering
\hfill %
\subfloat[][Keypoint Transfer Accuracy.]{
\begin{tabular}{@{}ll|rrrr@{}} \toprule
Supv & Method & Bird & Horse  & Cow & Sheep \\ \midrule
\multirow{2}{0.7cm}{KP + Mask}
 & CMR ~\cite{kanazawa2016learning}  & 47.3 & -- & -- & -- \\
 & CSM ~\cite{kulkarni2019canonical} & 45.8 & 42.1 & 28.5 & 31.5 \\
 & A-CSM ~\cite{kulkarni2020articulation} & 51.0  & 44.6 & 29.2 & 39.0 \\
 & IMR  (ours) &  {\bf 58.5} & {\bf  51.0} & {\bf  38.7} & {\bf  40.5} \\
\midrule
Mask
&  Dense-Equi ~\cite{thewlis2017unsupervised} & 33.5 & 23.3 & 20.9 & 19.6 \\
& CSM ~\cite{kulkarni2019canonical} & 36.4 & 31.2 & 26.3 & 24.7 \\
&  A-CSM ~\cite{kulkarni2020articulation} & 42.6 & 32.9 & 26.3 & 28.6 \\
&   IMR (ours) & {\bf 53.4} & {\bf  46.2} & {\bf  33.2} & {\bf  32.4} \\
\bottomrule
\end{tabular}
}
\qquad
\hfill %
\subfloat[][Keypoint Reprojection Accuracy.]{
\begin{tabular}{@{}ll|rrrrr@{}} \toprule
Supv & Method & Bird & Horse & Cow & Sheep\\ \midrule
\multirow{2}{7mm}{KP + Mask}
 & CMR ~\cite{kanazawa2016learning}  & 80.0 & --  & -- & -- \\
 & CSM ~\cite{kulkarni2019canonical}  & 68.5 & 46.4 & 52.6 & 47.9 \\
 & A-CSM ~\cite{kulkarni2020articulation} & 72.4 & 57.3 & 56.8 & {\bf 57.4} \\
 & IMR (ours) & {\bf  83.7} & {\bf 63.3} & {\bf 60.0} & 55.7  \\
\midrule
Mask
  & CSM ~\cite{kulkarni2019canonical} & 50.9 & 49.7 & 37.4 & 36.4 \\
  & A-CSM ~\cite{kulkarni2020articulation} & 46.8 & 54.2 & 43.8 & 42.5 \\
  & IMR  (ours) & {\bf 57.0} & {\bf 59.5} & {\bf 53.2} & {\bf 46.9}  \\
\bottomrule
\end{tabular}
}
\hfill
\vspace{2mm}
\caption{\textbf{Evaluation via Semantic Keypoints}. We report comparisons to previous approaches for: a) transferring semantic keypoints between images using the induced dense correspondence via our 3D predictions, and b) Predicting 2D keypoints via reprojection of 3D keypoints. Higher is better.}
\tablelabel{kpeval}
\end{table}

%% file: figures.tex
\begin{table*}[!t]
\setlength{\tabcolsep}{0.01em}
\renewcommand{\arraystretch}{1}
\centering
  \scalebox{0.63}{
\begin{tabular}{lccc@{\hskip 1em}lccc}
\addpic{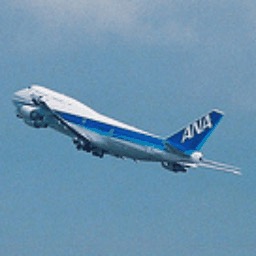} & 
\addpic{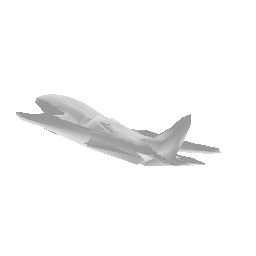} & 
\addpic{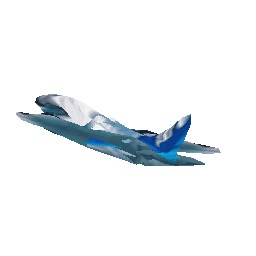}& 
\addpic{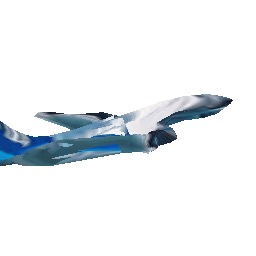}& 
\addpic{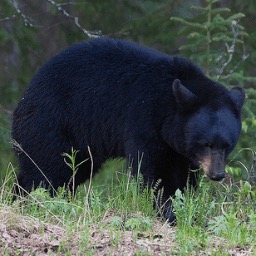} & 
\addpic{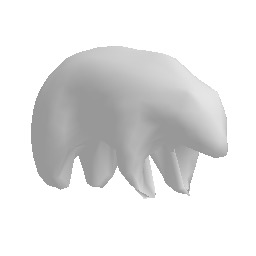} & 
\addpic{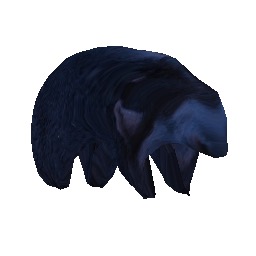}& 
\addpic{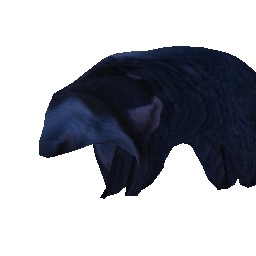} \\
\addpic{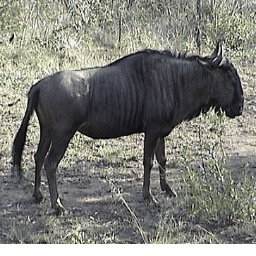} & 
\addpic{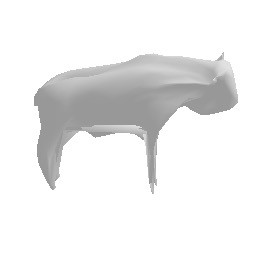} & 
\addpic{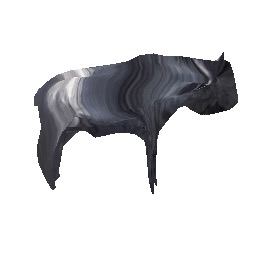}& 
\addpic{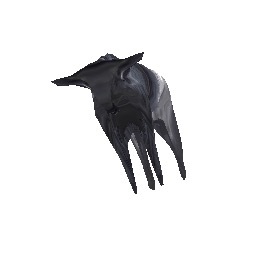}& 
\addpic{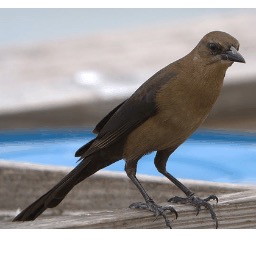} & 
\addpic{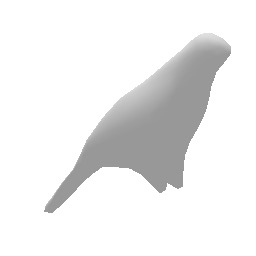} & 
\addpic{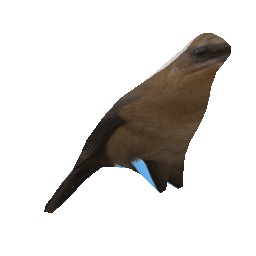}& 
\addpic{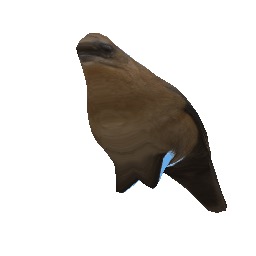} \\ 
\addpic{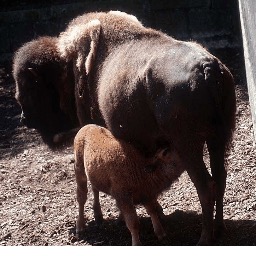} & 
\addpic{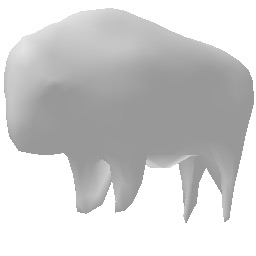} & 
\addpic{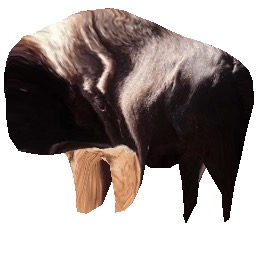}& 
\addpic{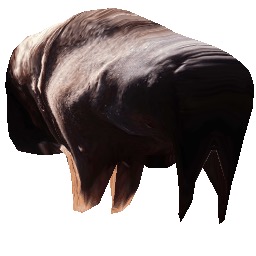}& 
\addpic{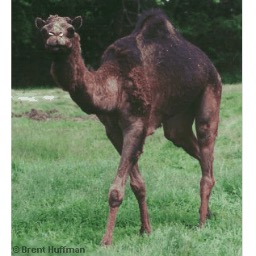} & 
\addpic{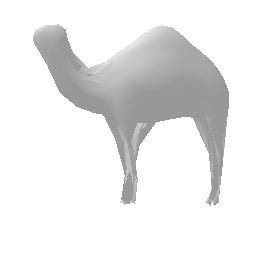} & 
\addpic{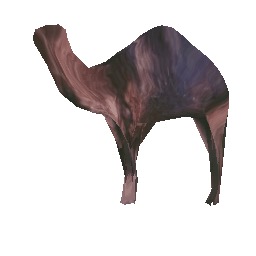}& 
\addpic{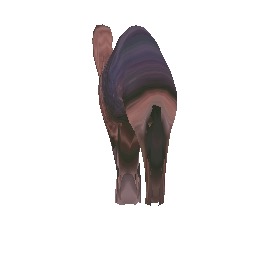} \\ 
\addpic{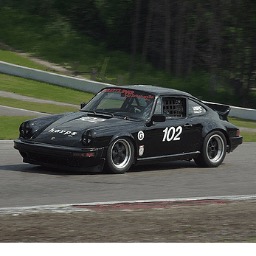} & 
\addpic{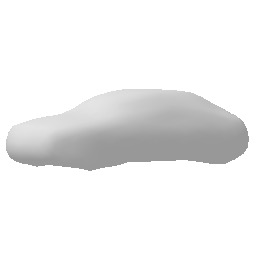} & 
\addpic{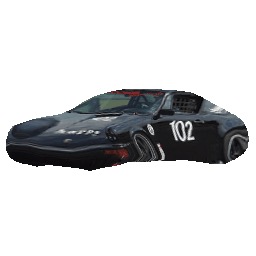}& 
\addpic{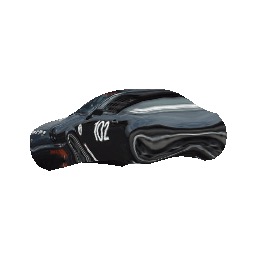}& 
\addpic{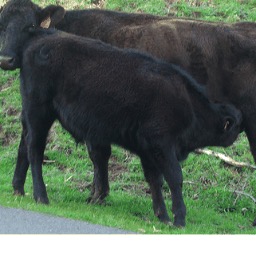} & 
\addpic{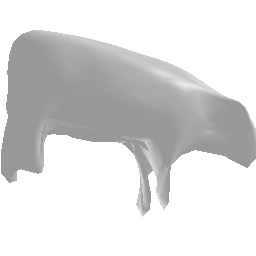} & 
\addpic{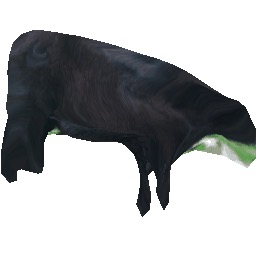}& 
\addpic{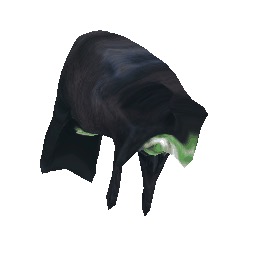} \\ 
\addpic{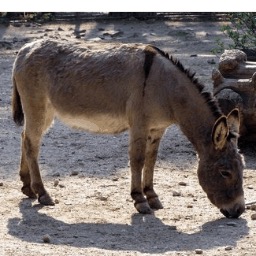} & 
\addpic{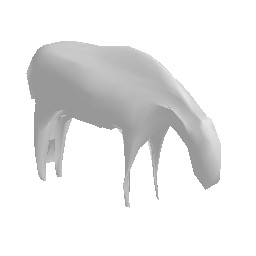} & 
\addpic{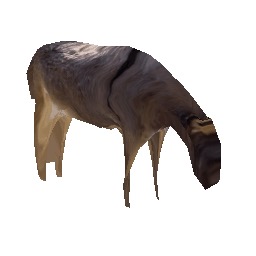}& 
\addpic{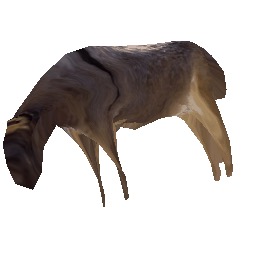}& 
\addpic{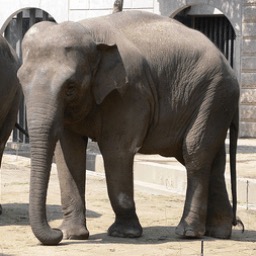} & 
\addpic{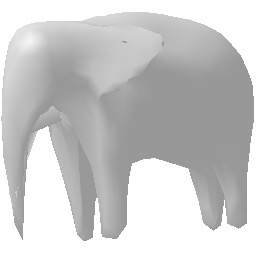} & 
\addpic{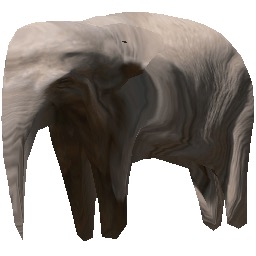}& 
\addpic{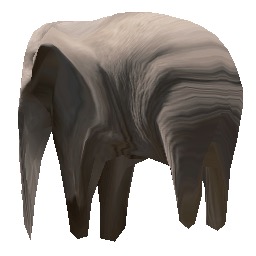} \\ 
\addpic{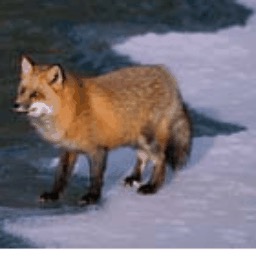} & 
\addpic{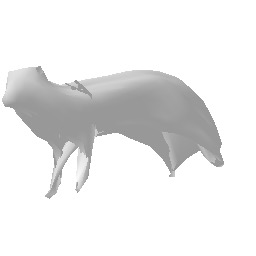} & 
\addpic{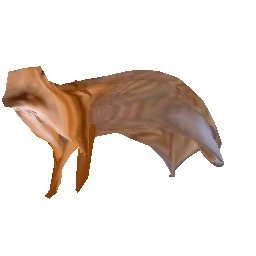}& 
\addpic{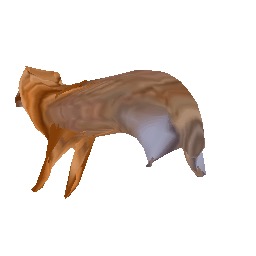}& 
\addpic{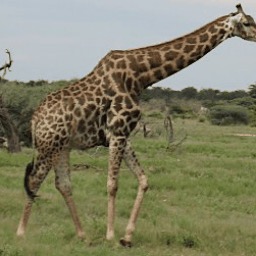} & 
\addpic{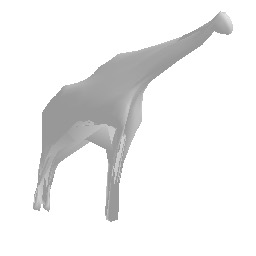} & 
\addpic{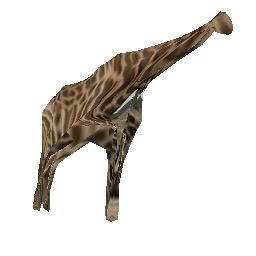}& 
\addpic{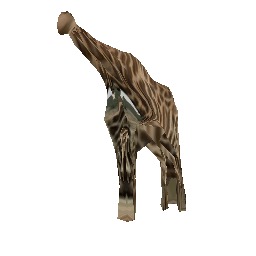} \\ 
\addpic{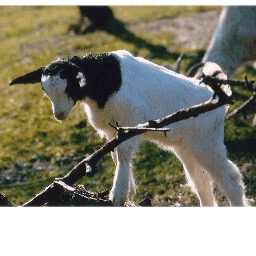} & 
\addpic{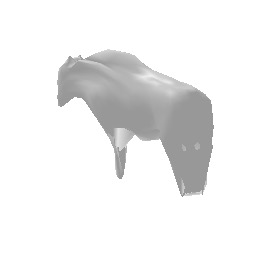} & 
\addpic{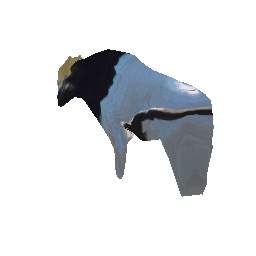}& 
\addpic{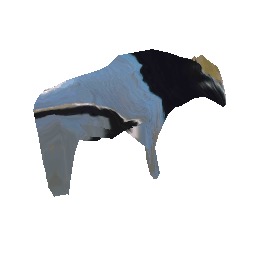}& 
\addpic{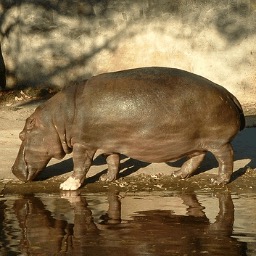} & 
\addpic{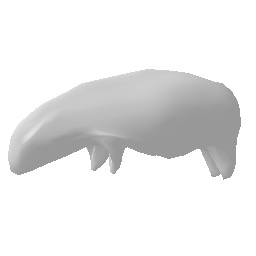} & 
\addpic{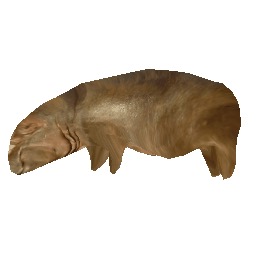}& 
\addpic{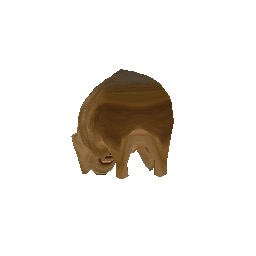} \\ 
\addpic{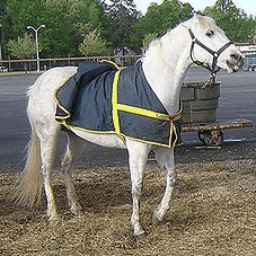} & 
\addpic{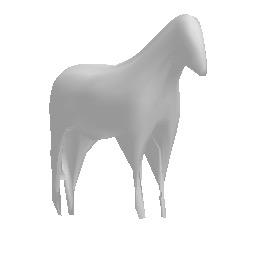} & 
\addpic{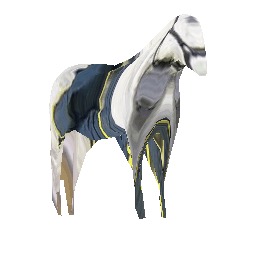}& 
\addpic{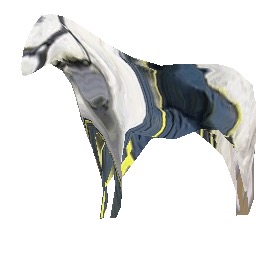}& 
\addpic{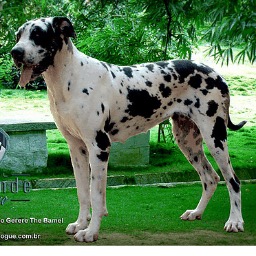} & 
\addpic{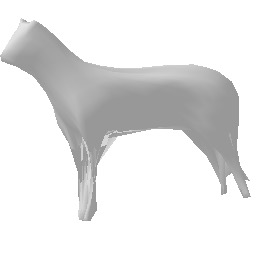} & 
\addpic{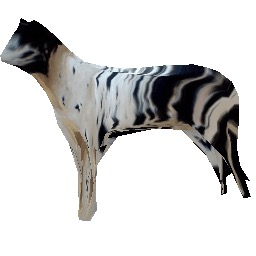}& 
\addpic{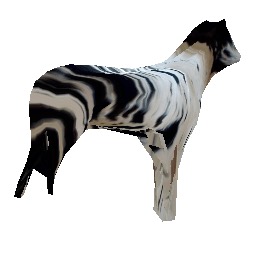}\\ 
\addpic{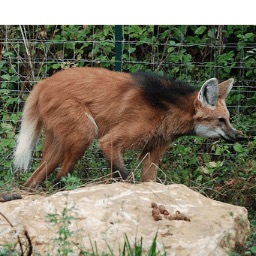} & 
\addpic{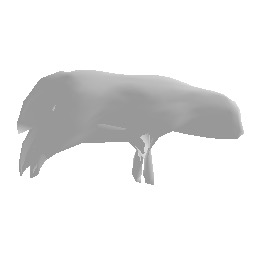} & 
\addpic{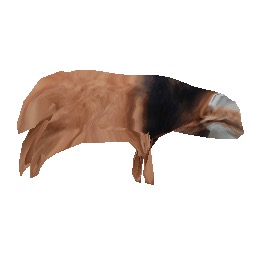}& 
\addpic{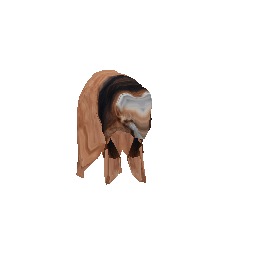}&
\addpic{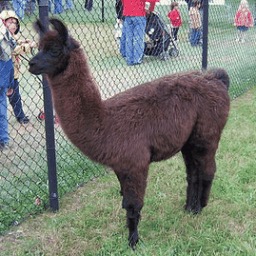} & 
\addpic{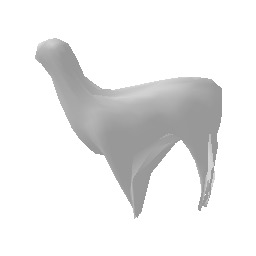} & 
\addpic{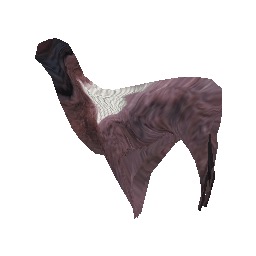}& 
\addpic{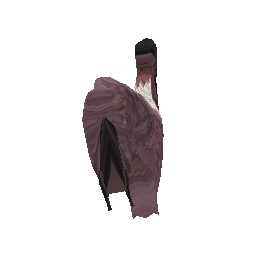} \\ 
\addpic{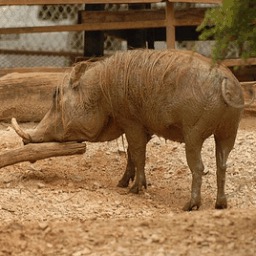} & 
\addpic{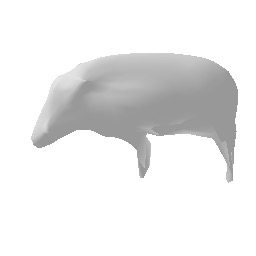} & 
\addpic{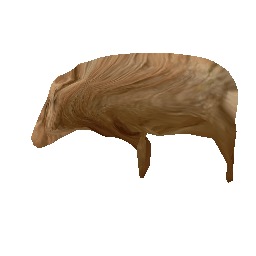}& 
\addpic{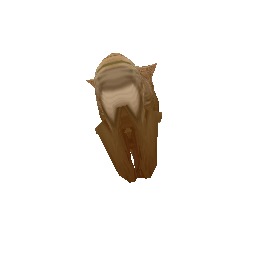}&
\addpic{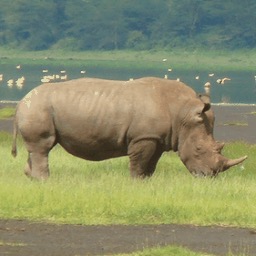} & 
\addpic{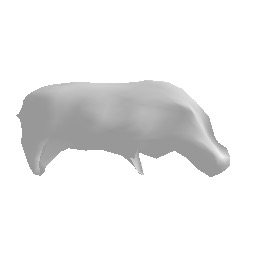} & 
\addpic{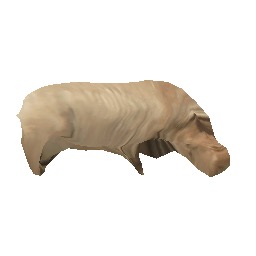}& 
\addpic{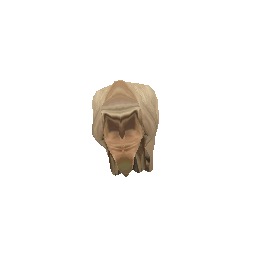} \\ 
\addpic{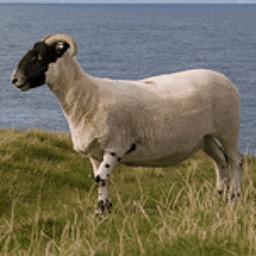} & 
\addpic{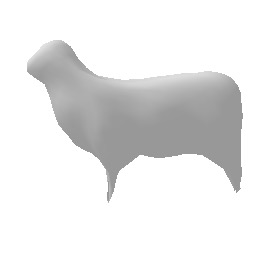} & 
\addpic{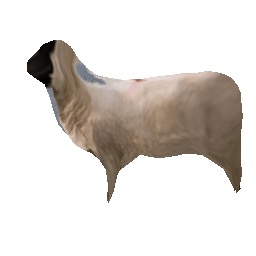}& 
\addpic{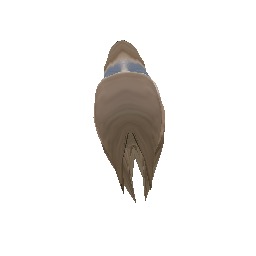}& 
\addpic{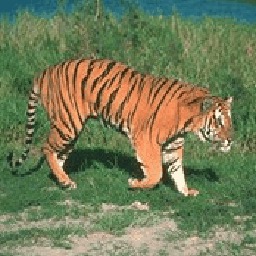} & 
\addpic{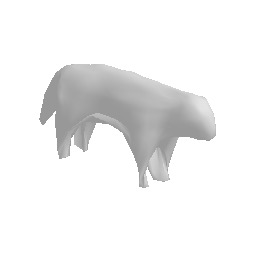} & 
\addpic{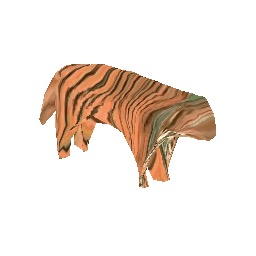}& 
\addpic{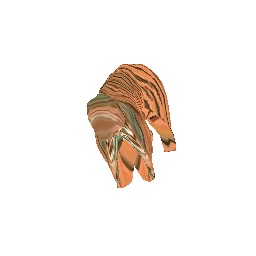}
\end{tabular}
}
\captionof{figure}{
\textbf{Sample Results.} We show (from left to right) the input image, the inferred 3D shape and texture from the predicted viewpoint, and the textured 3D shape from a novel viewpoint. Please see the \href{https://shubhtuls.github.io/imr/}{project page} and appendix for additional visualizations of results across all classes.
}
\figlabel{qual-results-detection}
\end{table*}

%% file: discussion.tex
\vspace{-2mm}
\section{Discussion}
\vspace{-2mm}
We presented an approach for learning implicit shape and texture inference from unannotated image collections. Although this enabled 3D prediction for a diverse set of categories, several challenges still remain towards being able to reconstruct thousands of object classes in generic images. As we model shapes via deformation of a (learned) template, our model does not allow for large or topological shape changes that maybe common in artificial object categories (\eg chairs). Additionally, our reprojection based objectives implicitly assume that the object is largely unoccluded, and it would be interesting to generalize these objectives to allow partial visibility. Lastly, our results do not always capture the fine details or precise pose, and it may be desirable to leverage additional learning signal if available \eg from videos. While there is clearly more progress needed to build systems that can reconstruct any object in any image, we believe our work represents an exciting step towards learning scalable and self-supervised 3D inference.


%% file: appendix.tex
\vspace{-2mm}
\section*{Appendix}
\vspace{-2mm}
\paragraph{Initializing Implicit Shape Space using a Template Mesh.}
To initialize the $\bar{\phi}(\cdot)$ in our category-level implicit shape representation, we use a single template shape per category. We do so by training the network $\bar{\phi}(\cdot)$ to minimize a matching loss with the available template. Specifically, denoting by $S$ the surface of a given template, we generate a target point set by randomly sampling $N=1000$ points on the surface: $P_t = \{p_i \sim S | i=1\cdots N\}$. We also randomly sample points on implicit shape corresponding to the current $\bar{\phi}(\cdot)$ via transforming random samples on the sphere $P_t = \{\bar{\phi}(u_i); u_i \sim \sps | i=1\cdots N\}$. We then iteratively train  $\bar{\phi}(\cdot)$ to minimize a hungarian matching loss between the two point clouds (using different random samples every iteration).

\begin{figure}[h!]
  \centering \includegraphics[width=1.0\textwidth]{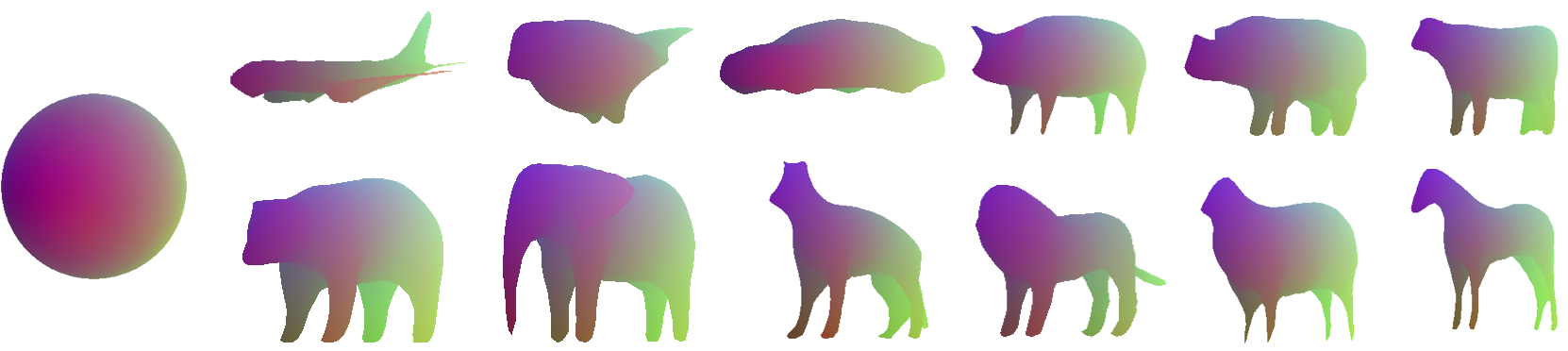}
  \caption{We visualize the obtained initialization for $\bar{\phi}$ for different categories. We use a single template shape per class, and learn $\bar{\phi}$ s.t. the shape represented via the deformation of a sphere matches the given template.}
  \figlabel{templates}
\end{figure}

\vspace{-2mm}
We visualize the resulting meshes learned for some categories in \figref{templates}, and and the sphere and the resulting shapes to highlight correspondence.  We observe that these capture the underlying 3D mesh well, but exhibit certain artifacts \eg hind legs of cow (top right). While this initialization helps in learning, we note that the learned shape space and inferred deformations allow us to model shapes that significantly vary from this initial template \eg fat vs thin bird, articulated animals. Please see the main text and additional visualizations below for examples.

\paragraph{Boundary Reprojection Consistency.}
To encourage our inferred 3D shapes to match the foreground mask boundary, we adapt the objectives proposed by Kar \etal~\cite{kar2015category}. Specifically, we encourage that the projected 3D points should lie inside the object, and that each point on the 2D boundary of the foreground mask should be close to some projected point(s).

Let us denote by $I_{fg}$ the foreground mask image corresponding to RGB image $I$ (from which we predicted a shape $\mathbf{z}_s$ and camera $\pi$). Further, let $\mathcal{B}_{fg}$ represent 2D points on the mask boundary, and $\mathcal{D}_{fg}$ correspond to a distance field induced by the mask. Additionally, let $P = \{\bar{\phi}(u_i); u_i \sim \sps | i=1\cdots N\}$ represent a set of $N$ randomly sampled points on the predicted 3D shape. Using these notations, our boundary reprojection objective can be formulated as:

\begin{gather*}
    L_{\text{boundary}} = \underset{p \in P}{\mathop{{}\mathbb{E}}}~ \mathcal{D}_{fg}(\pi(p)) ~~~+~~~ \underset{b \in \mathcal{B}_{fg}}{\mathop{{}\mathbb{E}}}~ \min_{p \in P}\|\pi(p)-b\|
\end{gather*}

\begin{wrapfigure}{r}{0.4\textwidth}
  \begin{center}
\vspace{-2.5em}
    \includegraphics[width=0.35\textwidth]{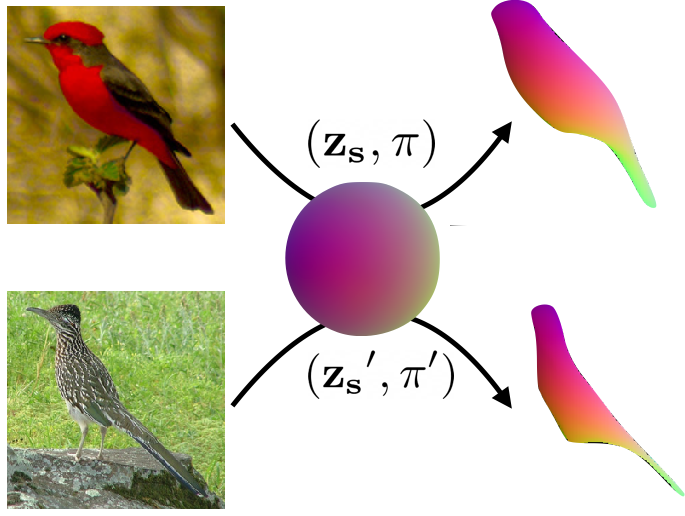}
  \end{center}
  \caption{Our inferred 3D representation allow us to render a per-pixel spherical coordinate, which can then be used to transfer semantics across two different images.}
  \figlabel{transfer}
\end{wrapfigure}
\paragraph{Correspondence Transfer via 3D Inference.} As our category-specific implicit shape  representation deforms a common sphere to yield the shape for any given instance, each mesh point is associated with a unique spherical coordinate. Given a predicted shape $\mathbf{z_s}$ and pose $\pi$ for an image $I$, we can therefore render a per-pixel spherical coordinate (akin to rendering a textured mesh) as shown in \figref{transfer}. These per-pixel spherical mappings subsequently allow us to transfer semantics (\eg keypoints) from a source image to a target image. For example, given a keypoint annotation in a source image, we can use to rendered spherical coordinate at that (or nearest) pixel as a query to find the corresponding point in a target image.

%% file: figures_supp.tex
\begin{table*}[!t]
\setlength{\tabcolsep}{0.01em}
\renewcommand{\arraystretch}{1}
\centering
  \scalebox{0.63}{
\begin{tabular}{c@{\hskip 1em}c@{\hskip 1em}c}
\addpicsup{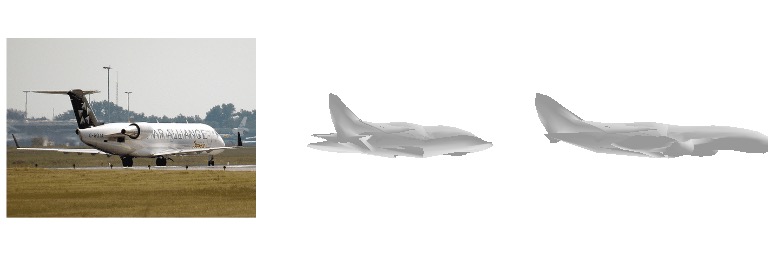} &
\addpicsup{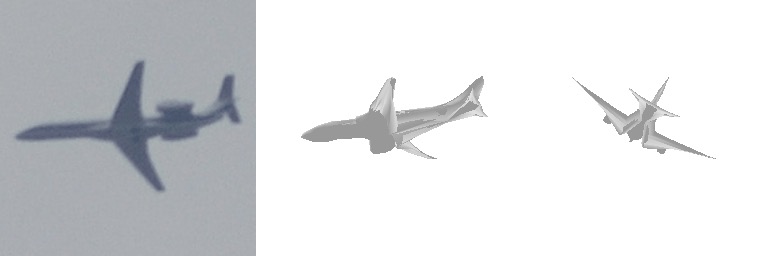} &
\addpicsup{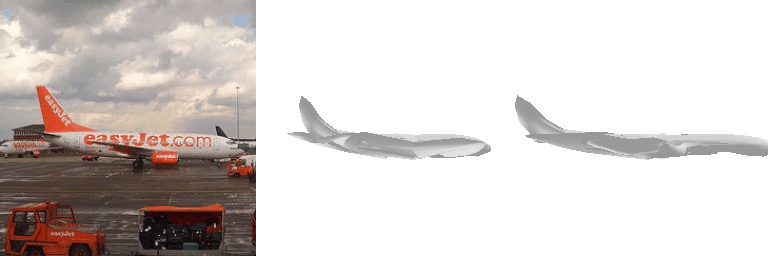} \\
\addpicsup{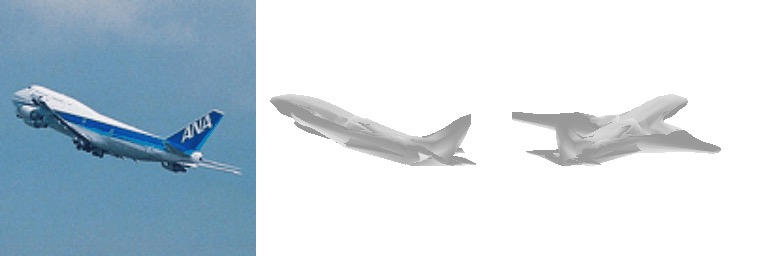} &
\addpicsup{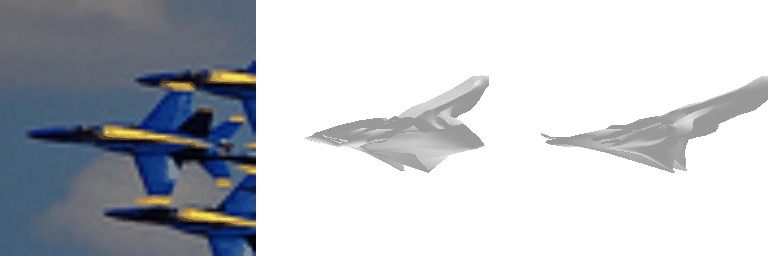} &
\addpicsup{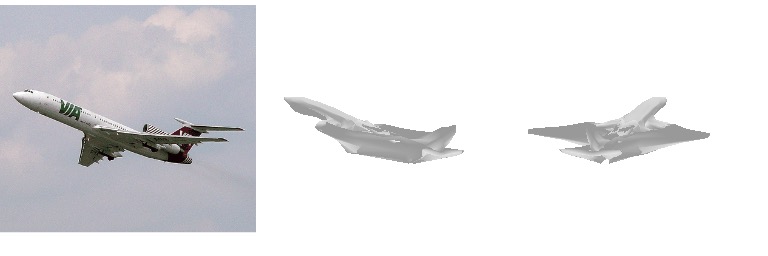} \\
\addpicsup{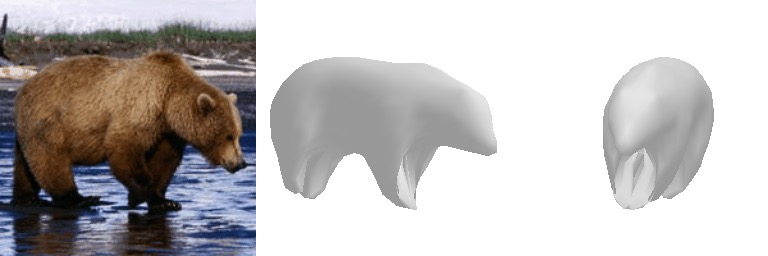} &
\addpicsup{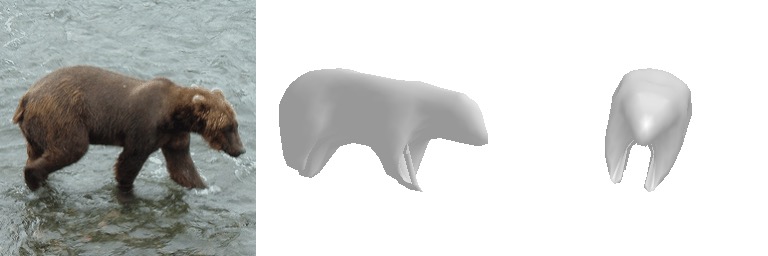} &
\addpicsup{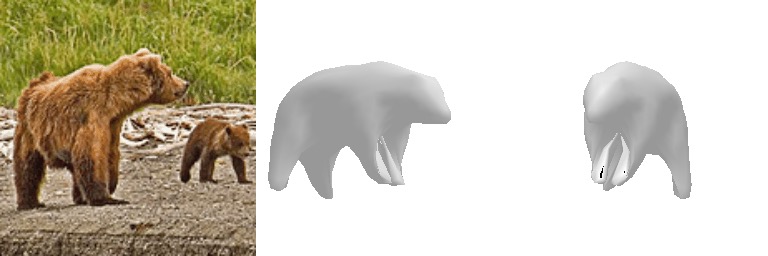} \\
\addpicsup{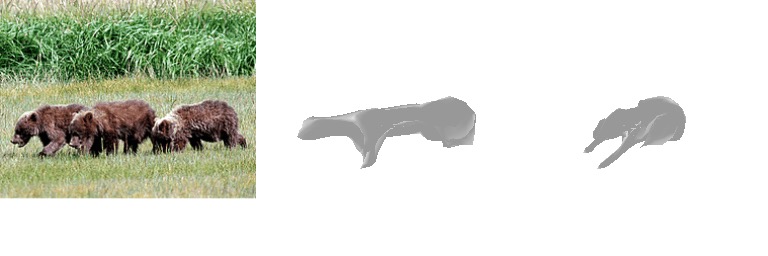} &
\addpicsup{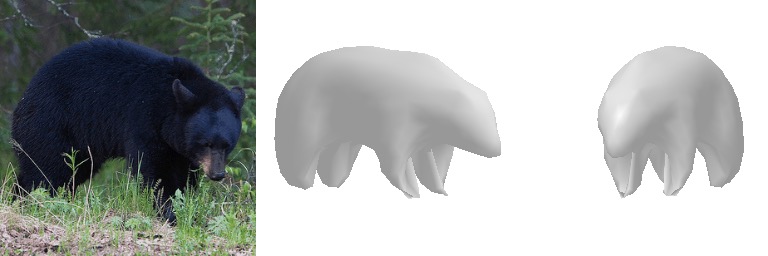} &
\addpicsup{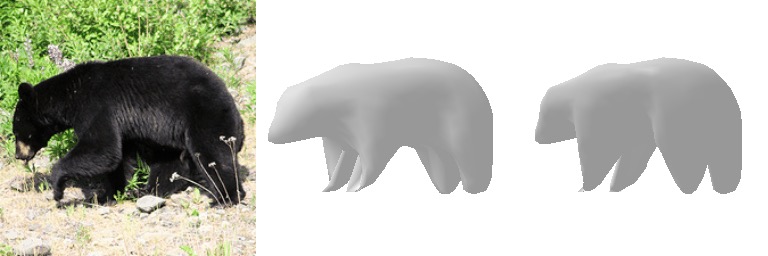} \\
\addpicsup{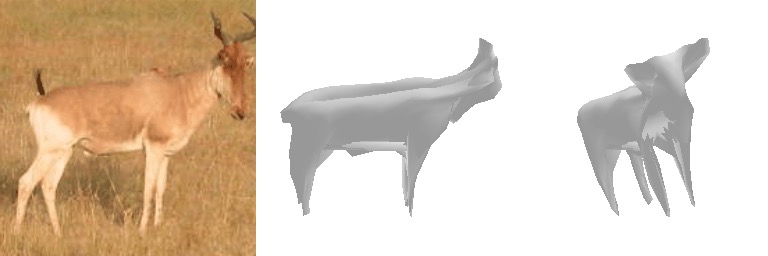} &
\addpicsup{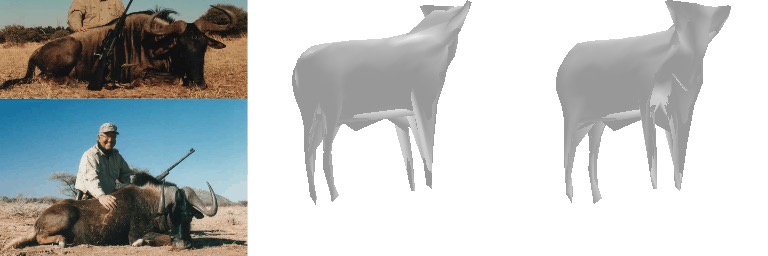} &
\addpicsup{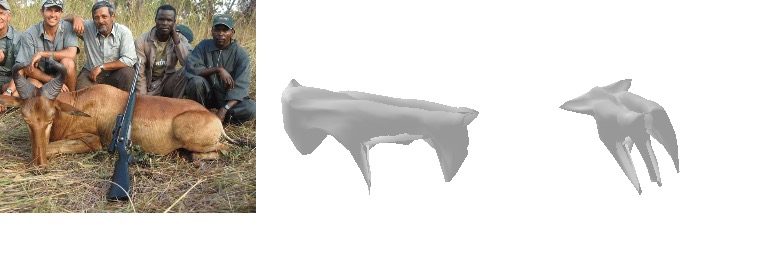} \\
\addpicsup{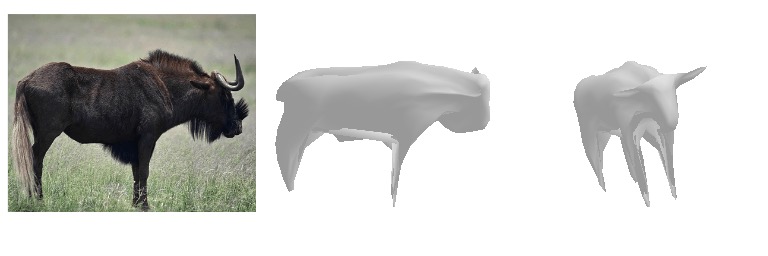} &
\addpicsup{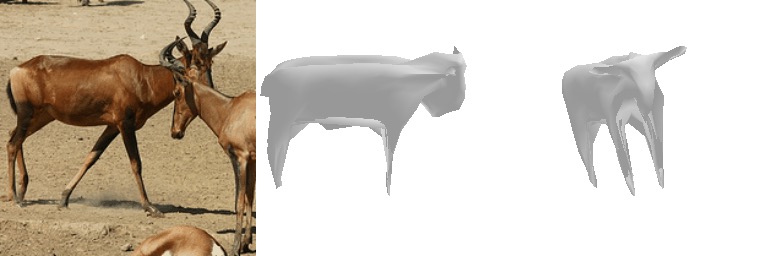} &
\addpicsup{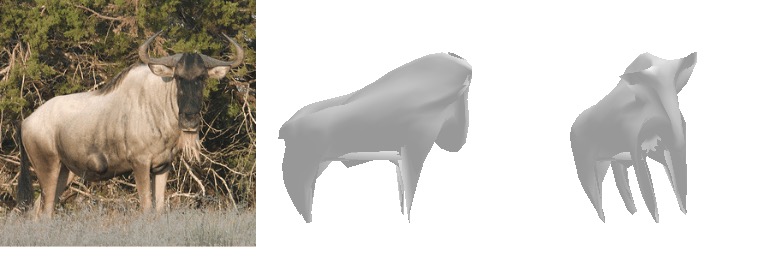} \\
\addpicsup{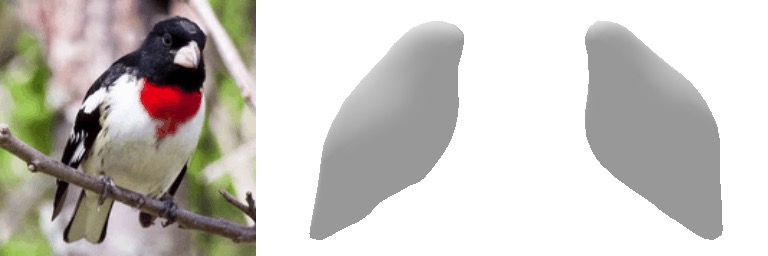} &
\addpicsup{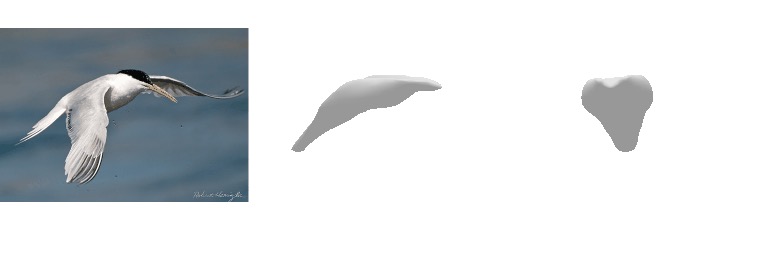} &
\addpicsup{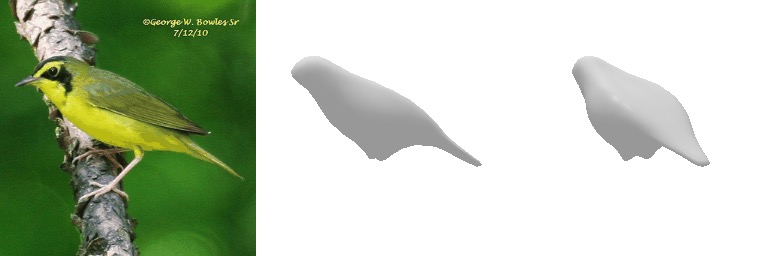} \\
\addpicsup{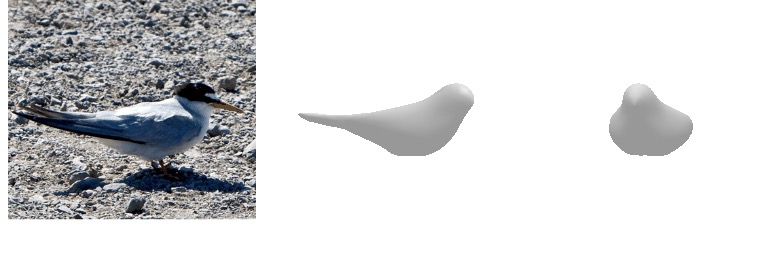} &
\addpicsup{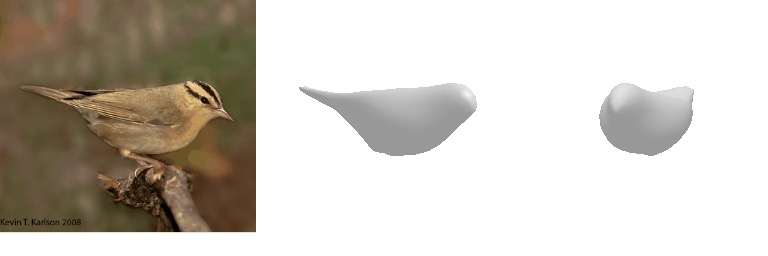} &
\addpicsup{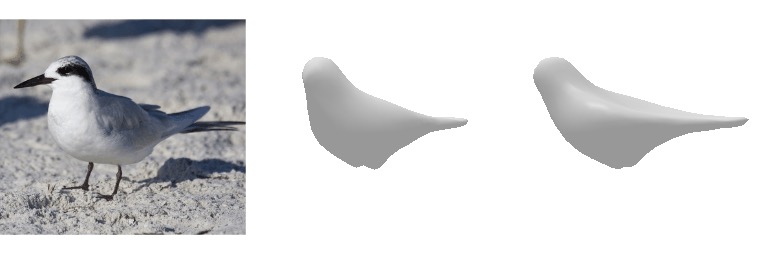} \\
\addpicsup{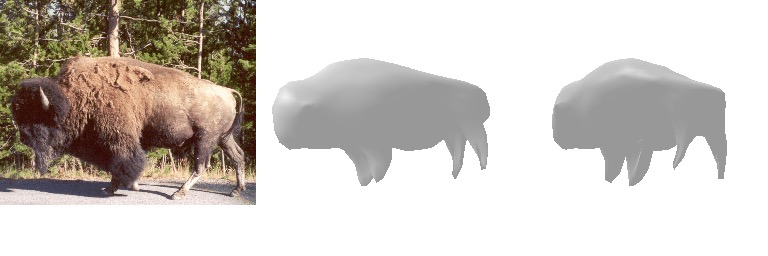} &
\addpicsup{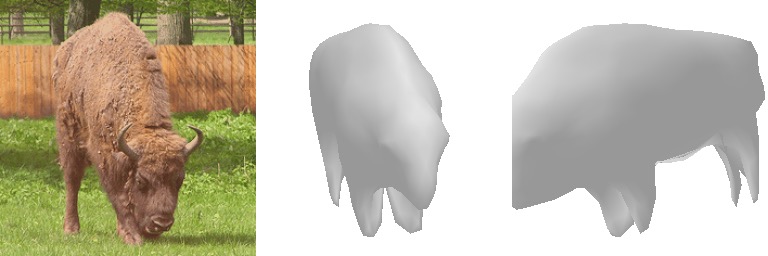} &
\addpicsup{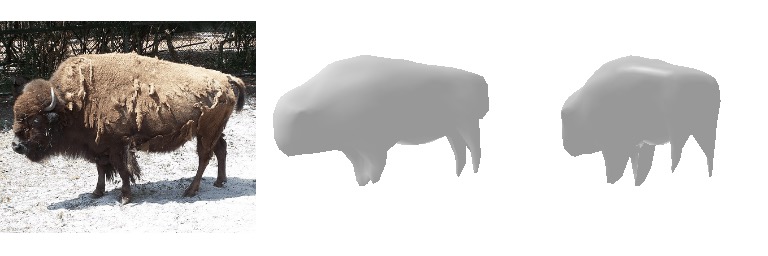} \\
\addpicsup{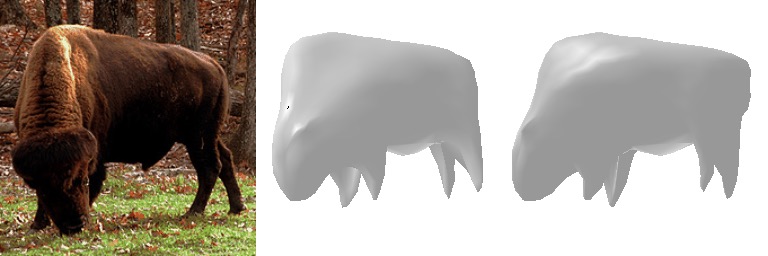} &
\addpicsup{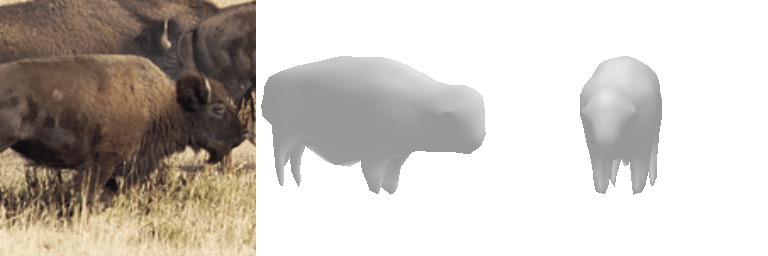} &
\addpicsup{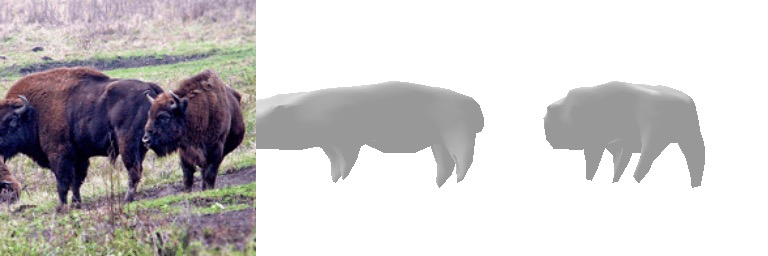} \\
\addpicsup{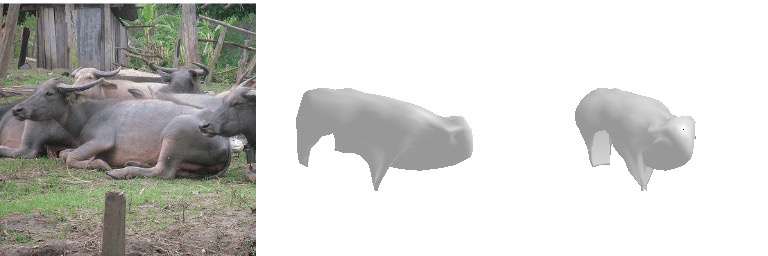} &
\addpicsup{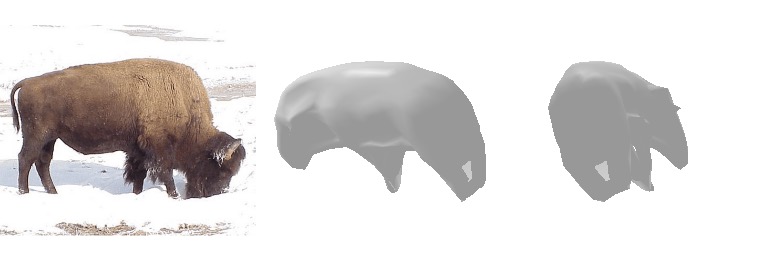} &
\addpicsup{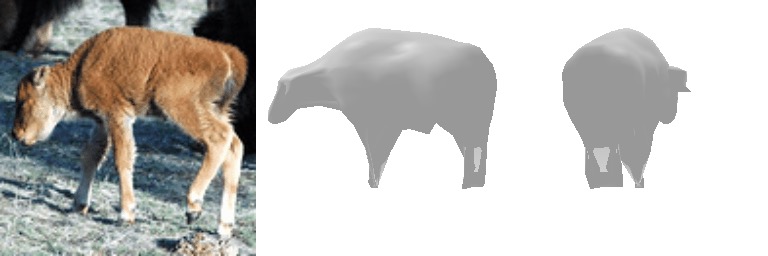} \\
\addpicsup{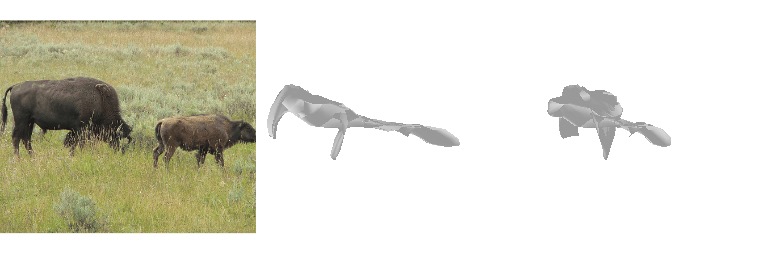} &
\addpicsup{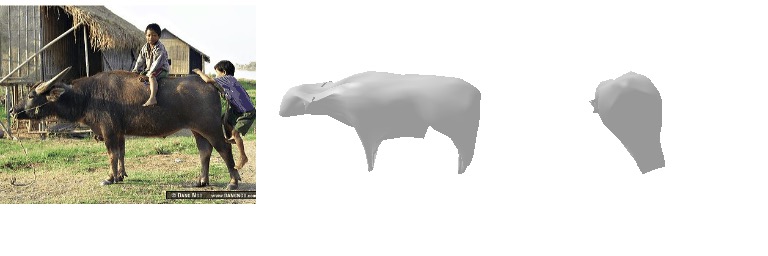} &
\addpicsup{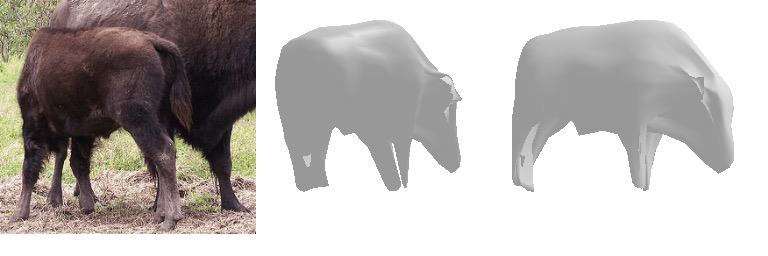} \\
\end{tabular}
}
\captionof{figure}{
\textbf{Random Results.} We show (from left to right) the input image, the inferred 3D shape from predicted view and a novel view for 6 \emph{randomly sampled} images from the test set per category.
}
\end{table*}

\begin{table*}[!t]
\setlength{\tabcolsep}{0.01em}
\renewcommand{\arraystretch}{1}
\centering
  \scalebox{0.63}{
\begin{tabular}{c@{\hskip 1em}c@{\hskip 1em}c}
\addpicsup{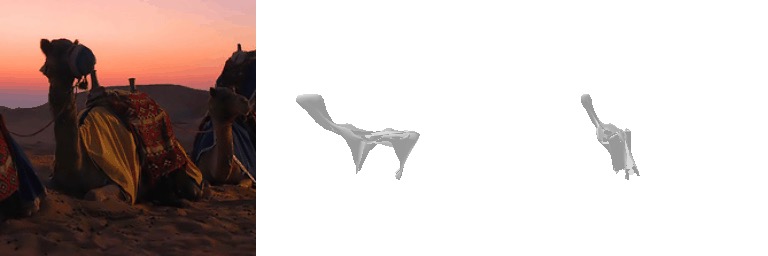} &
\addpicsup{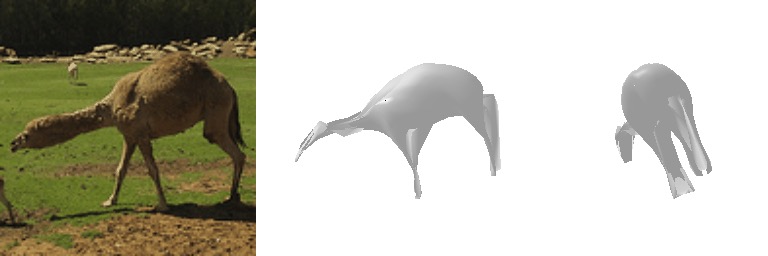} &
\addpicsup{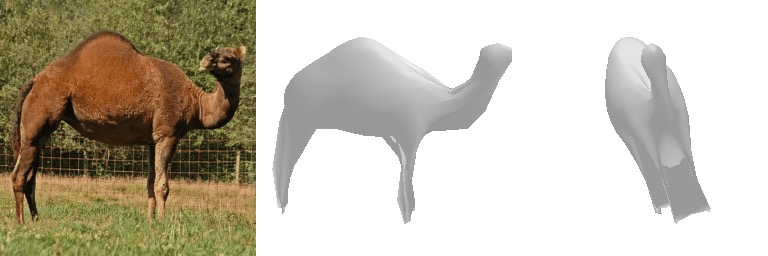} \\
\addpicsup{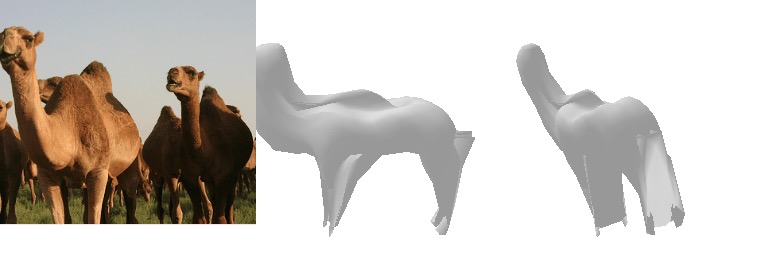} &
\addpicsup{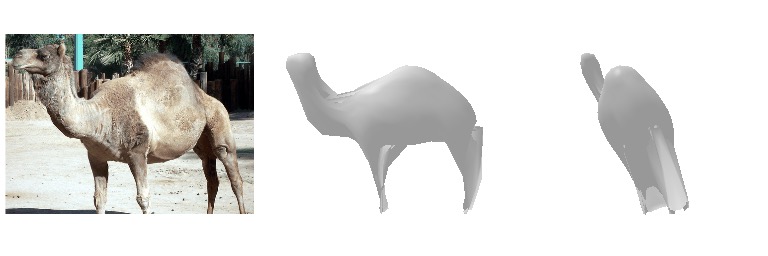} &
\addpicsup{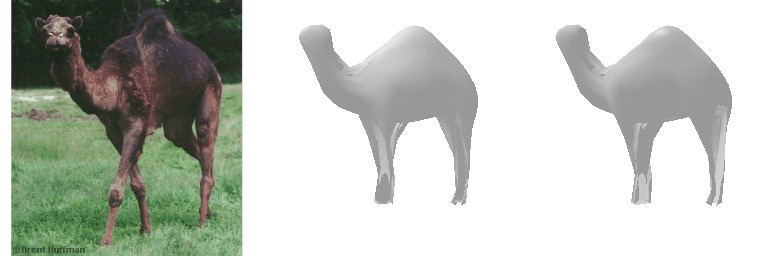} \\
\addpicsup{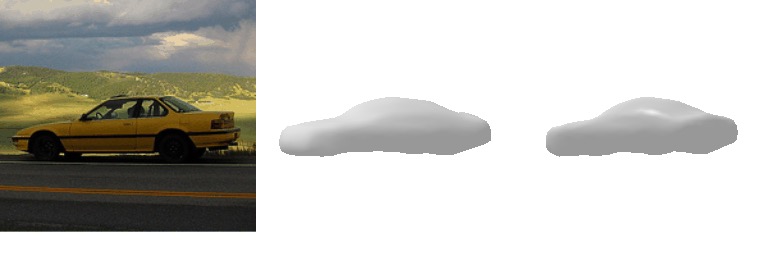} &
\addpicsup{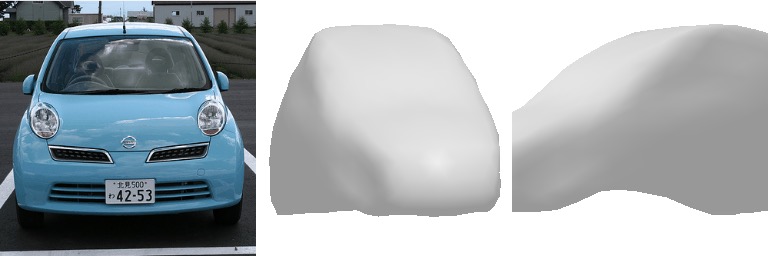} &
\addpicsup{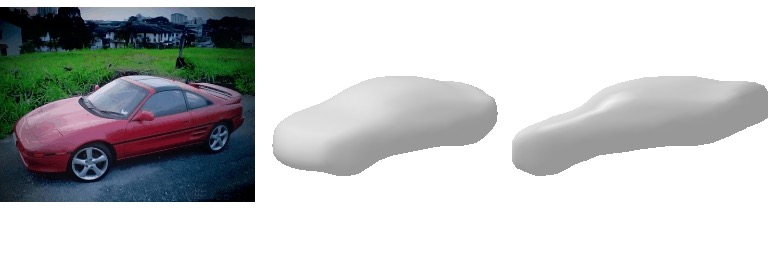} \\
\addpicsup{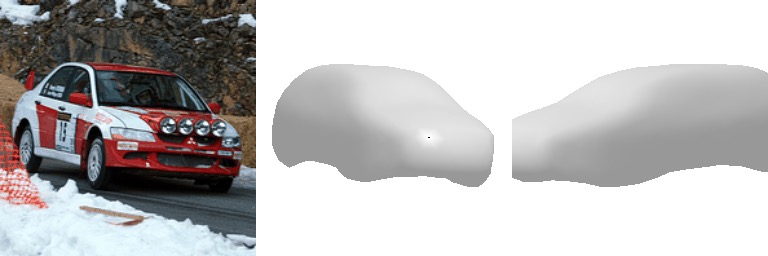} &
\addpicsup{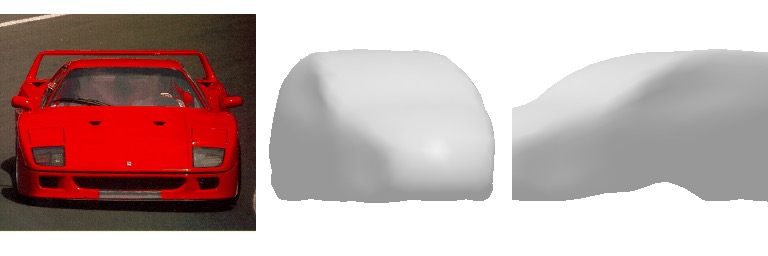} &
\addpicsup{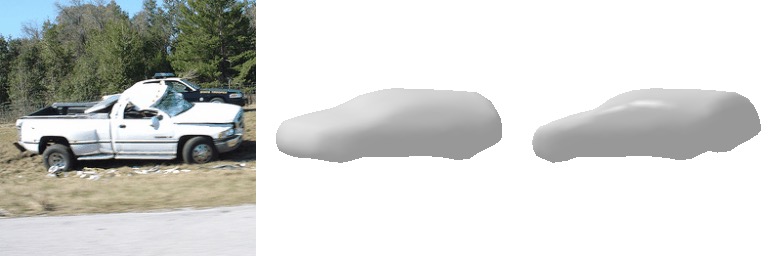} \\
\addpicsup{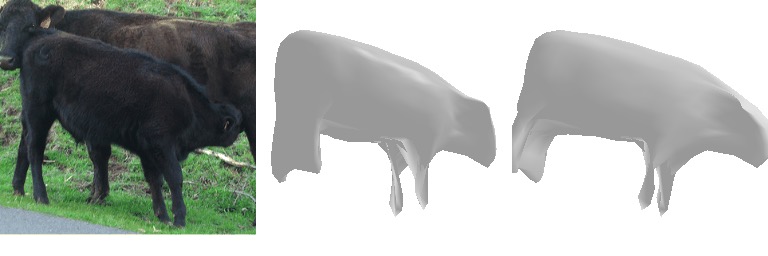} &
\addpicsup{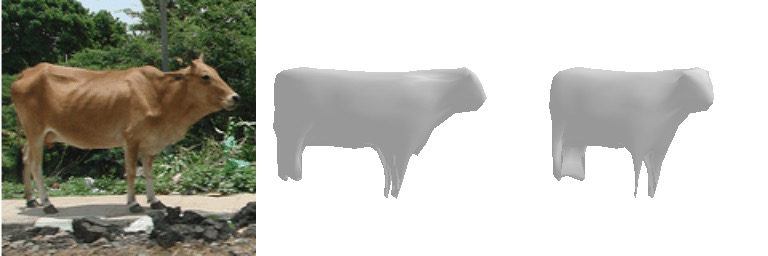} &
\addpicsup{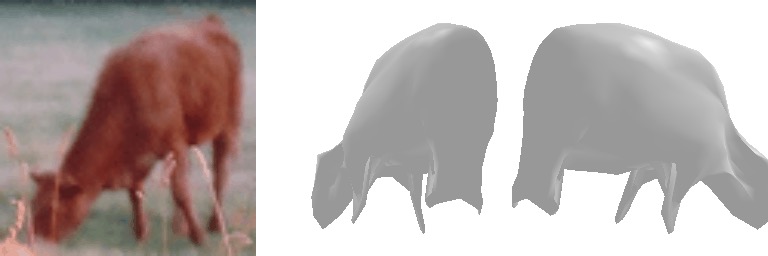} \\
\addpicsup{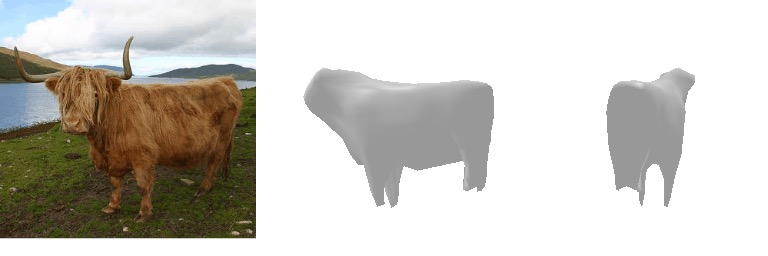} &
\addpicsup{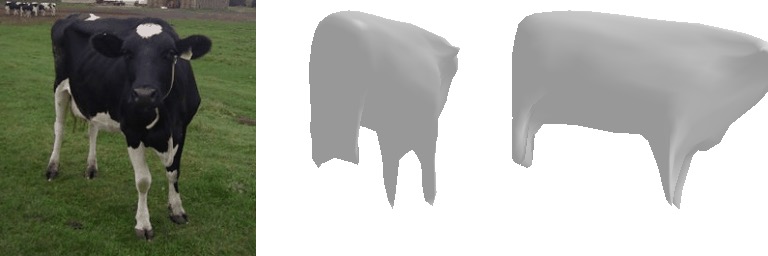} &
\addpicsup{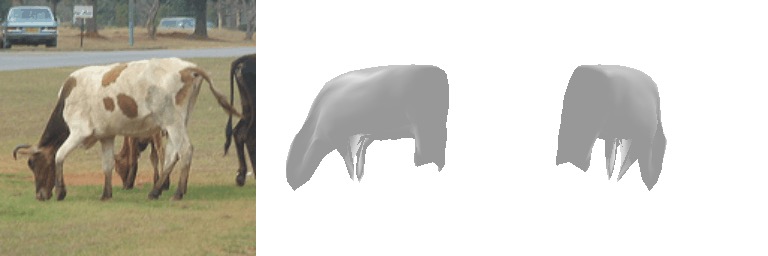} \\
\addpicsup{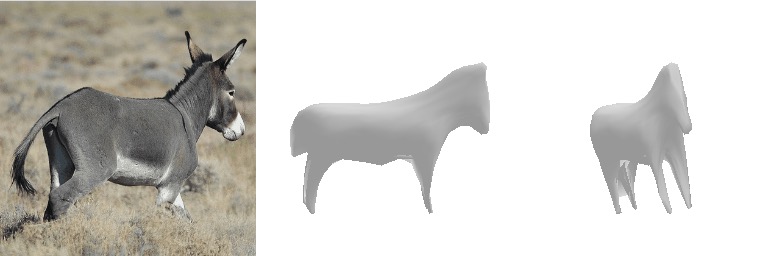} &
\addpicsup{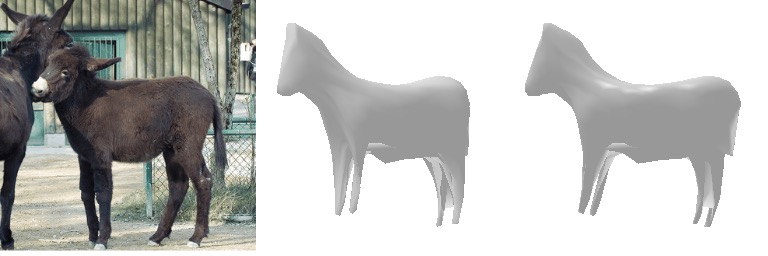} &
\addpicsup{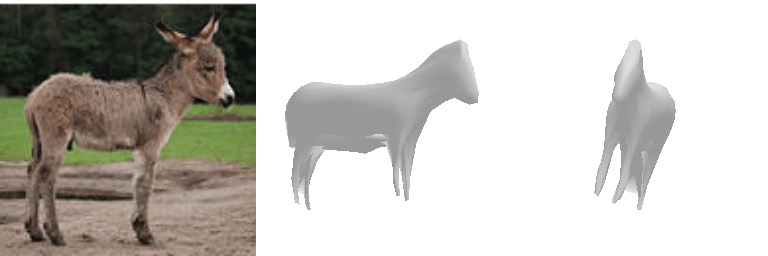} \\
\addpicsup{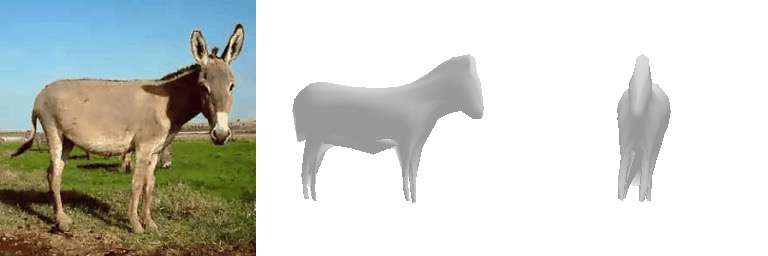} &
\addpicsup{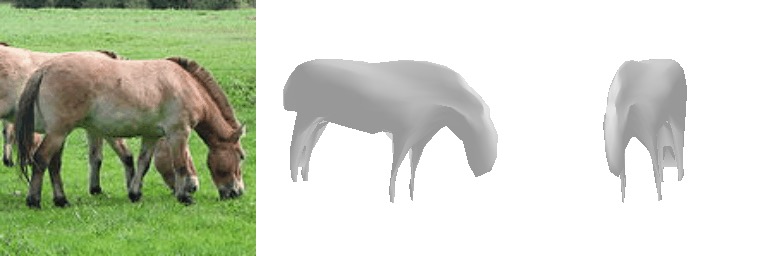} &
\addpicsup{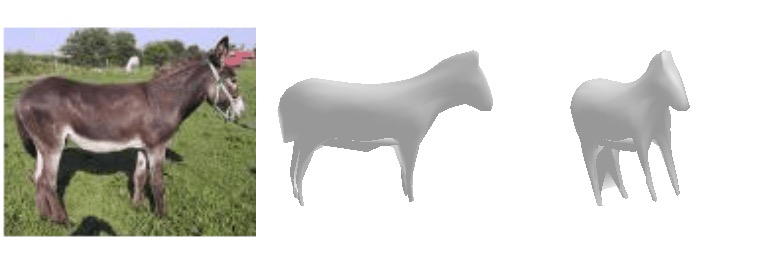} \\
\addpicsup{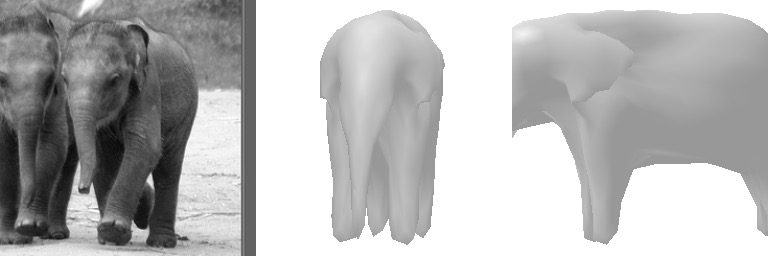} &
\addpicsup{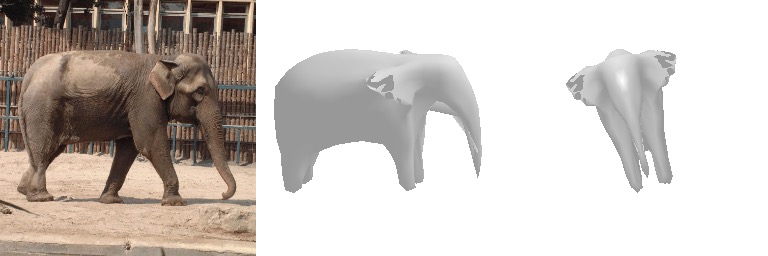} &
\addpicsup{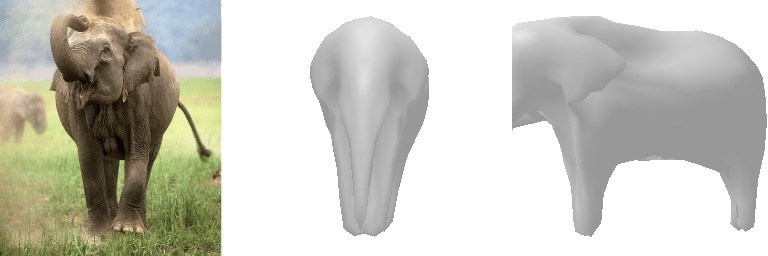} \\
\addpicsup{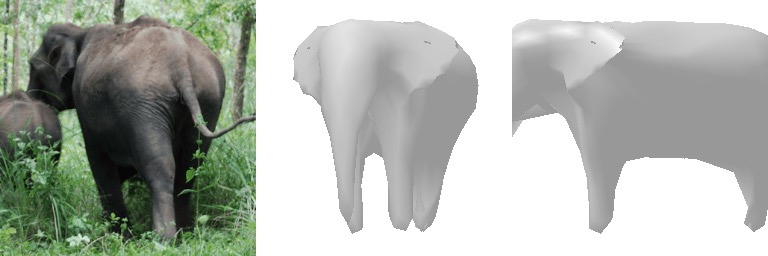} &
\addpicsup{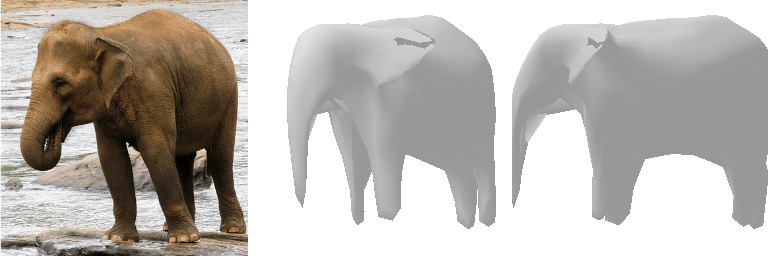} &
\addpicsup{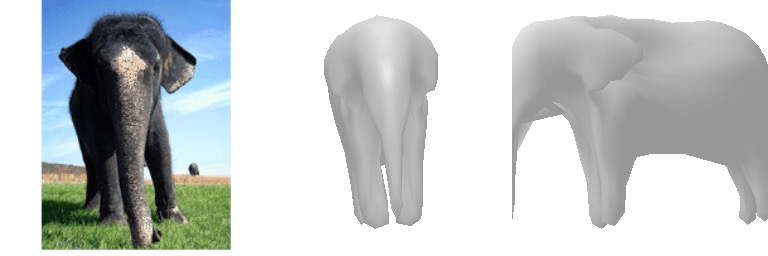} \\
\addpicsup{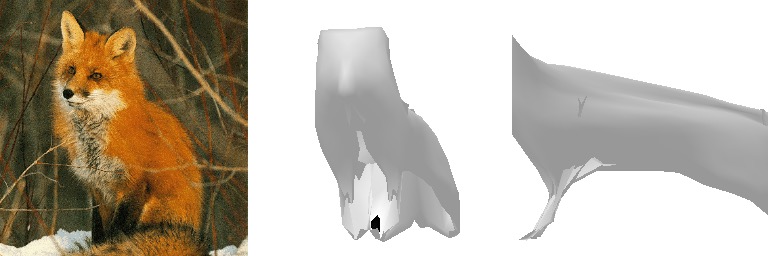} &
\addpicsup{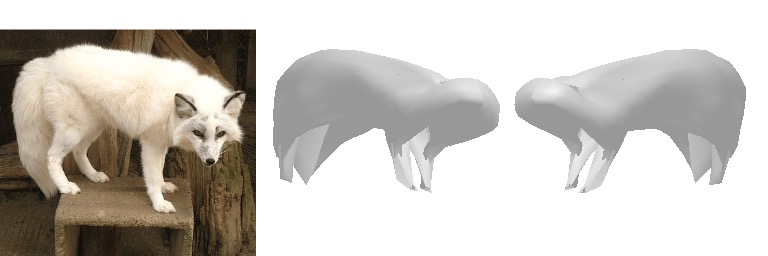} &
\addpicsup{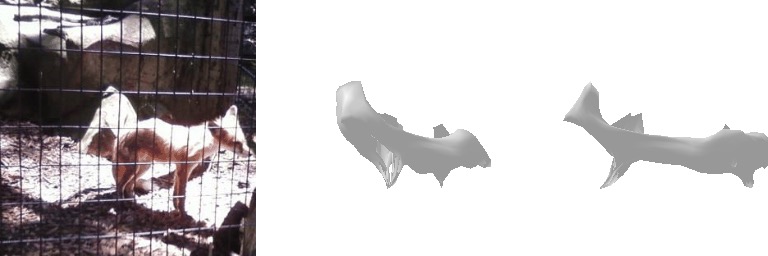} \\
\addpicsup{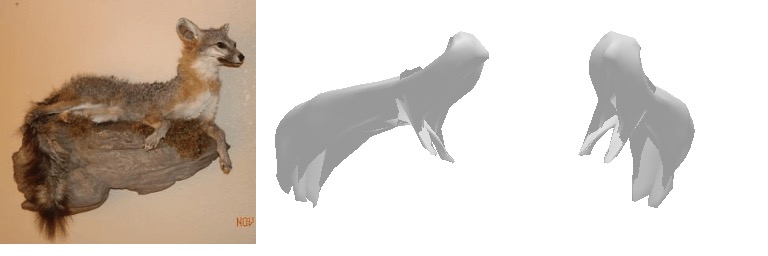} &
\addpicsup{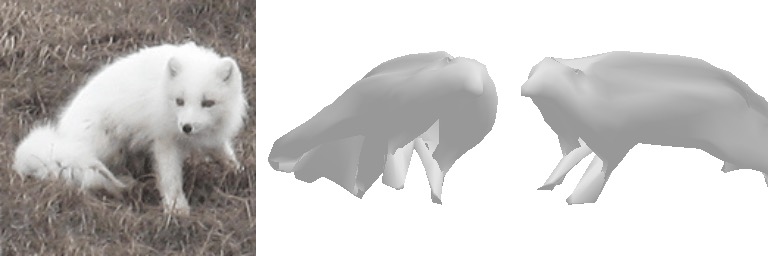} &
\addpicsup{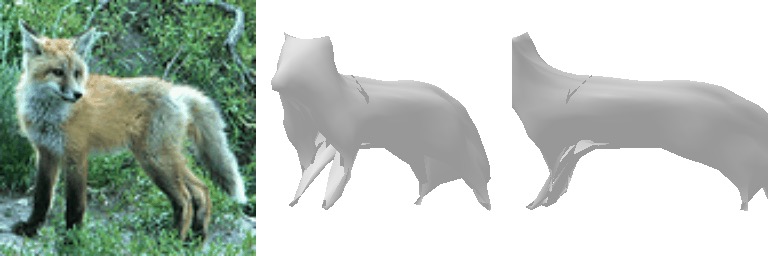} \\
\end{tabular}
}
\captionof{figure}{
\textbf{Random Results.} We show (from left to right) the input image, the inferred 3D shape from predicted view and a novel view for 6 \emph{randomly sampled} images from the test set per category.
}
\end{table*}

\begin{table*}[!t]
\setlength{\tabcolsep}{0.01em}
\renewcommand{\arraystretch}{1}
\centering
  \scalebox{0.63}{
\begin{tabular}{c@{\hskip 1em}c@{\hskip 1em}c}
\addpicsup{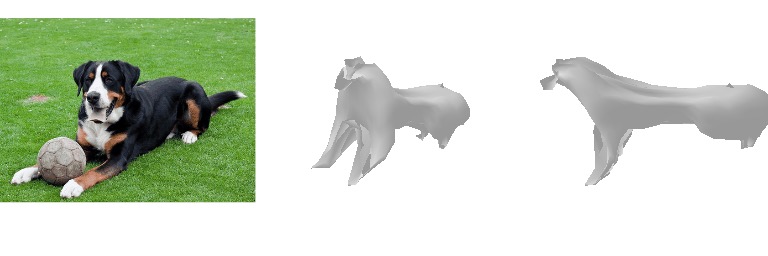} &
\addpicsup{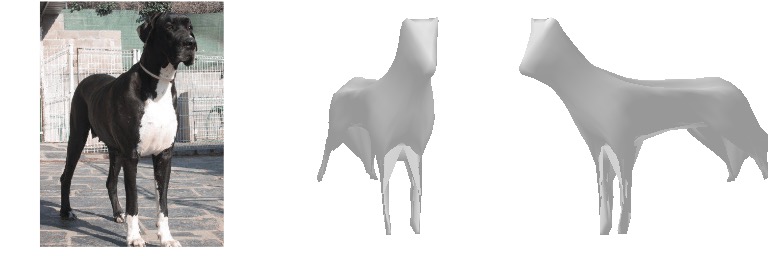} &
\addpicsup{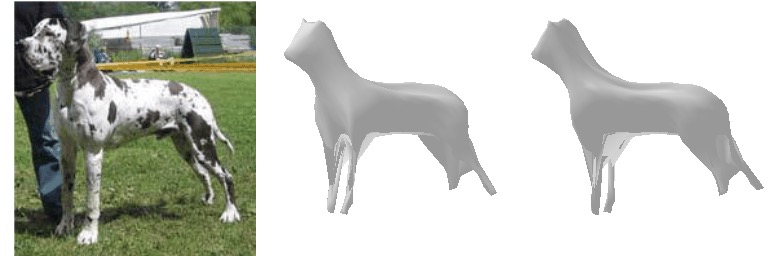} \\
\addpicsup{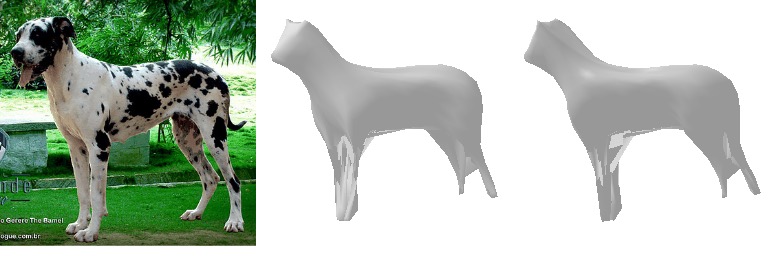} &
\addpicsup{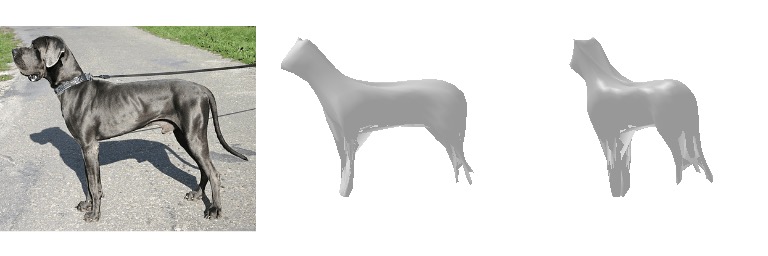} &
\addpicsup{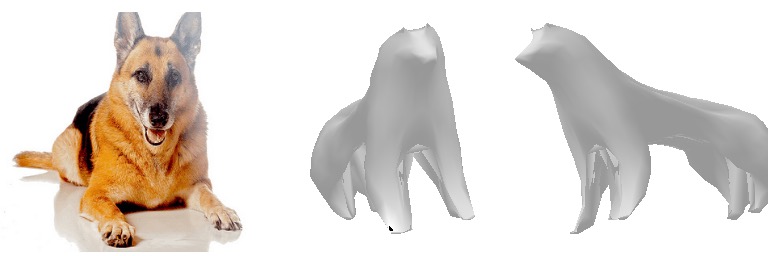} \\
\addpicsup{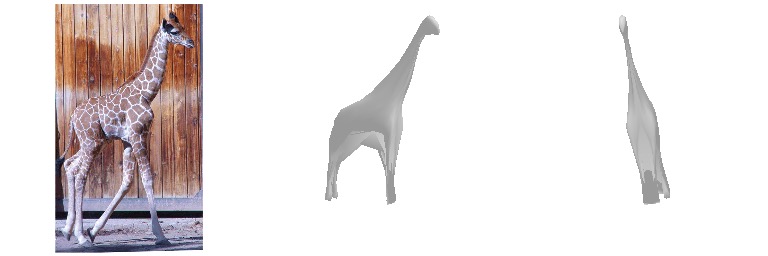} &
\addpicsup{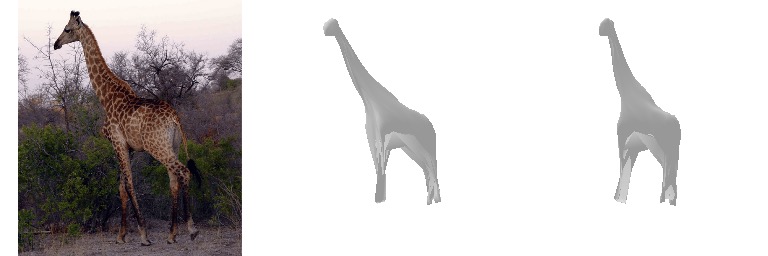} &
\addpicsup{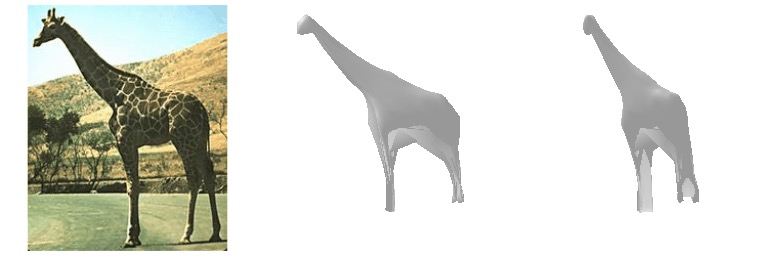} \\
\addpicsup{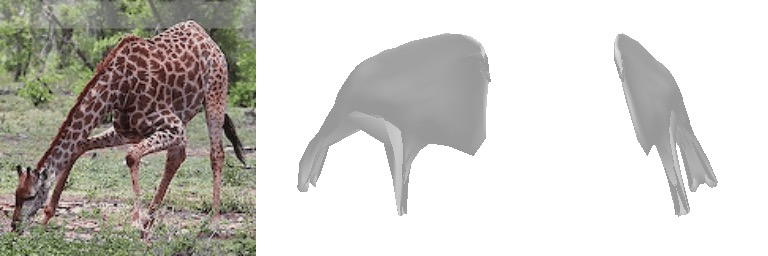} &
\addpicsup{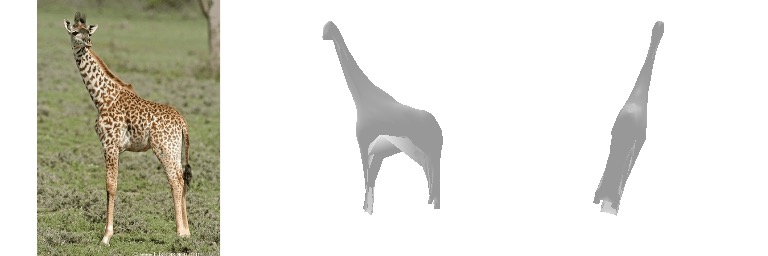} &
\addpicsup{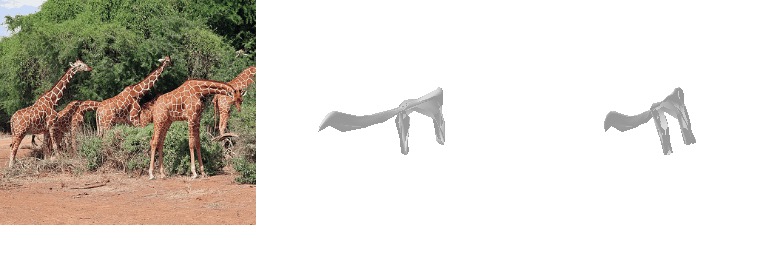} \\
\addpicsup{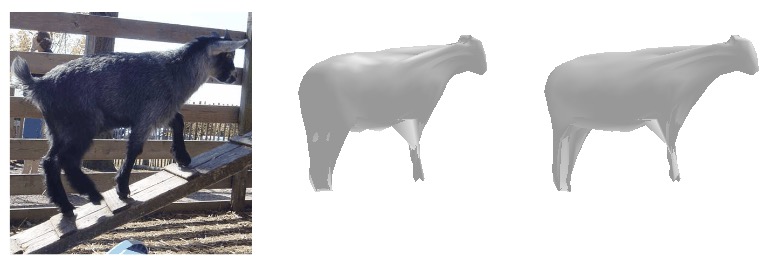} &
\addpicsup{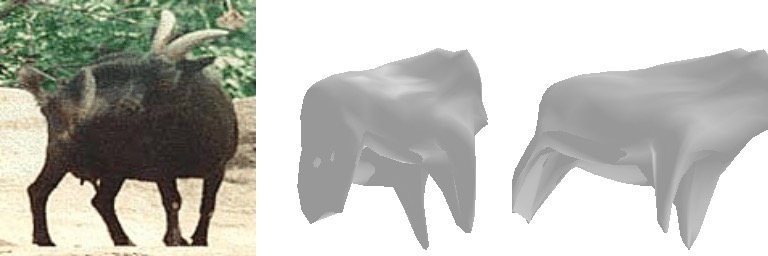} &
\addpicsup{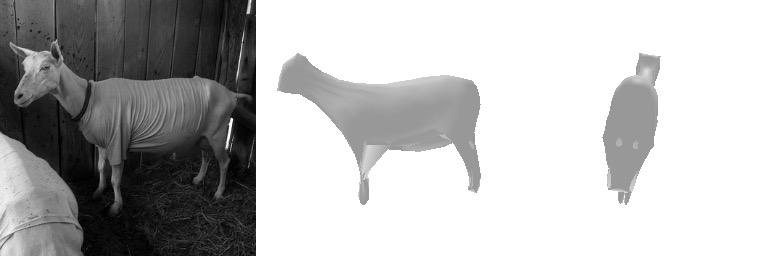} \\
\addpicsup{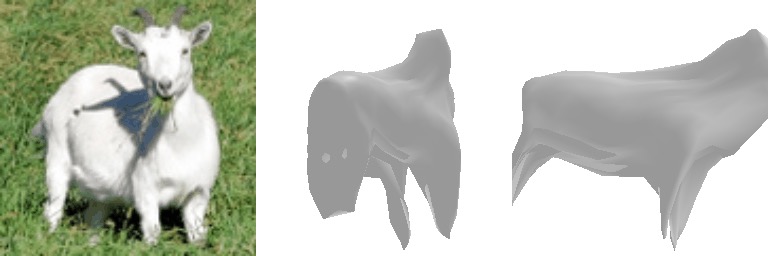} &
\addpicsup{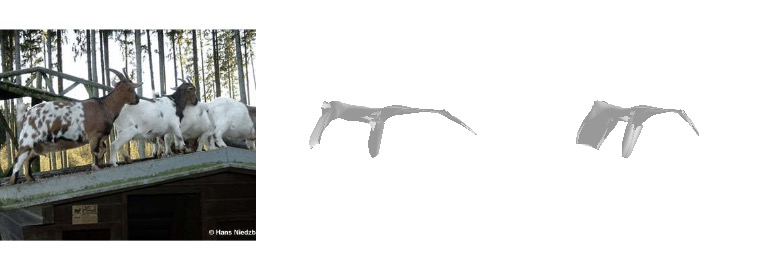} &
\addpicsup{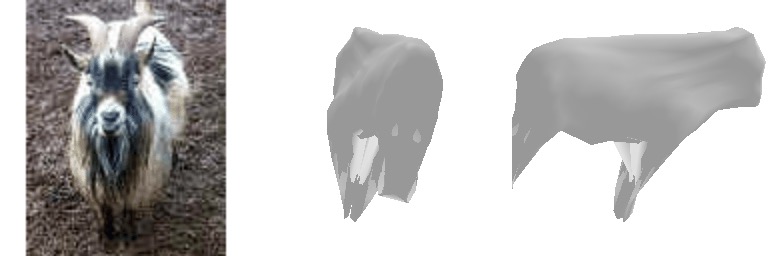} \\
\addpicsup{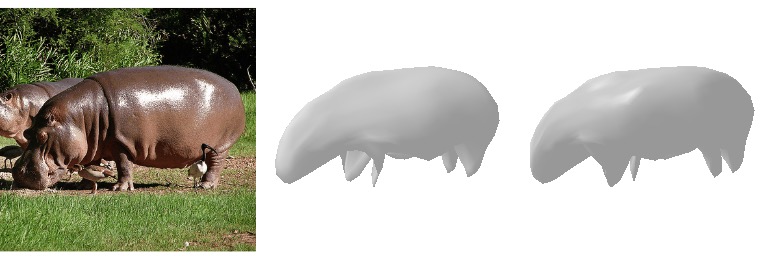} &
\addpicsup{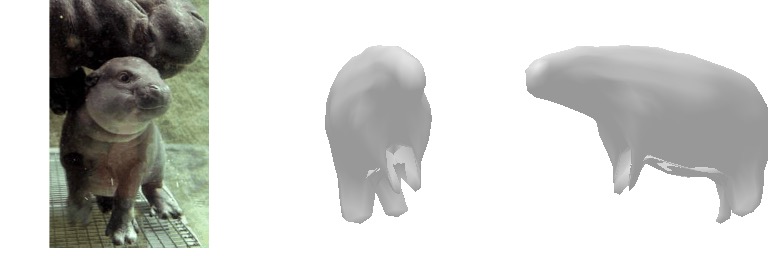} &
\addpicsup{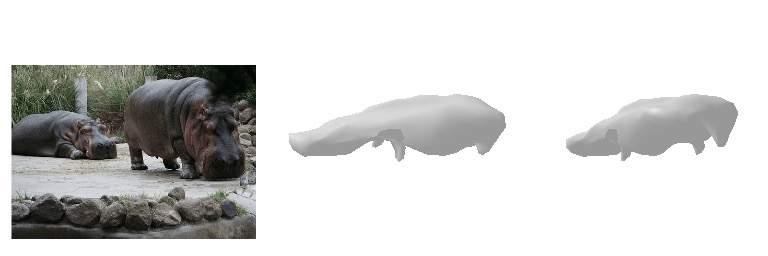} \\
\addpicsup{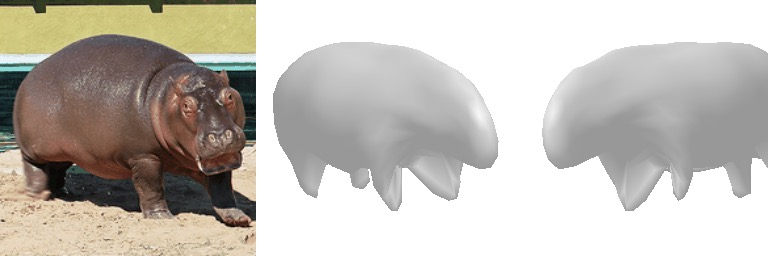} &
\addpicsup{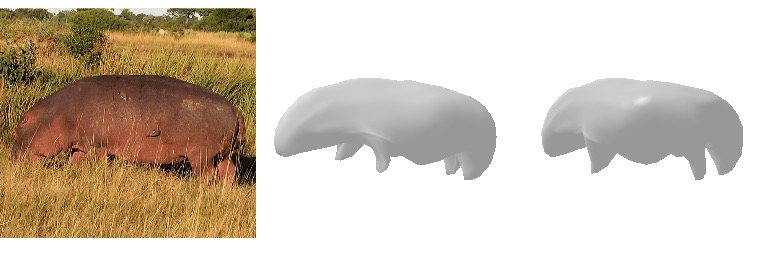} &
\addpicsup{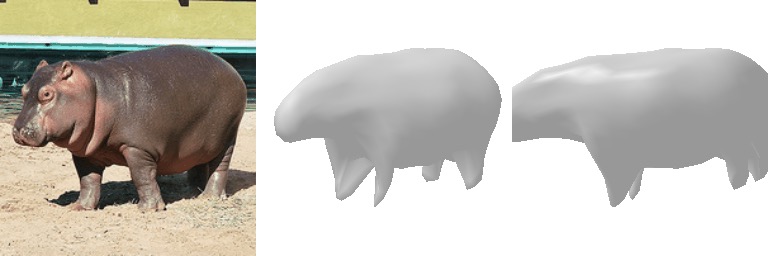} \\
\addpicsup{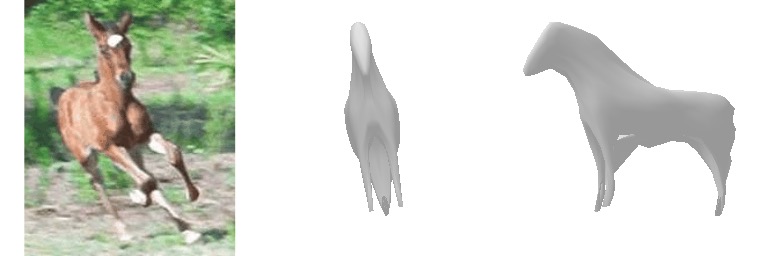} &
\addpicsup{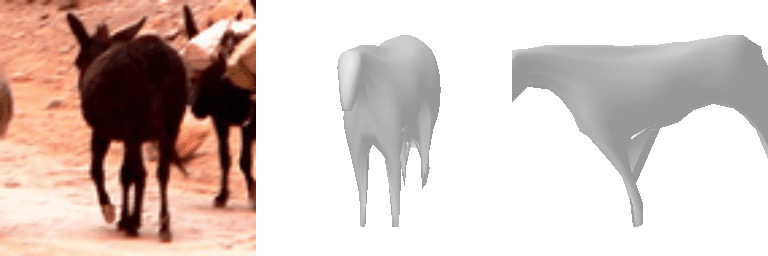} &
\addpicsup{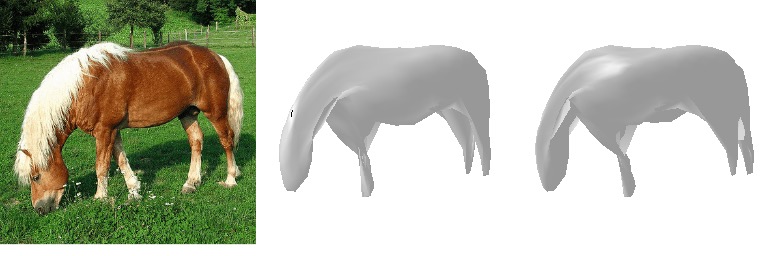} \\
\addpicsup{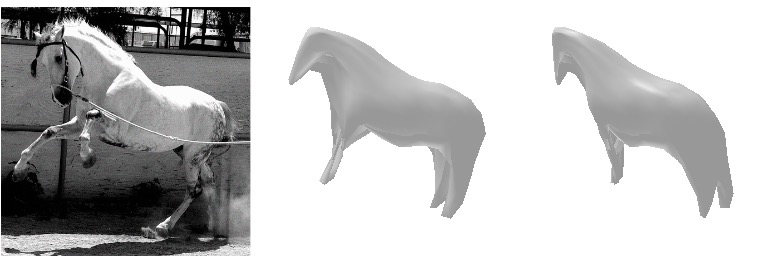} &
\addpicsup{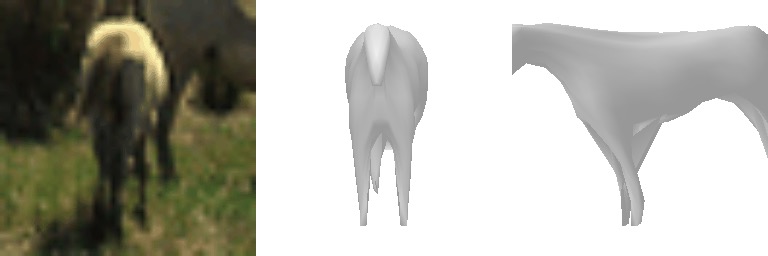} &
\addpicsup{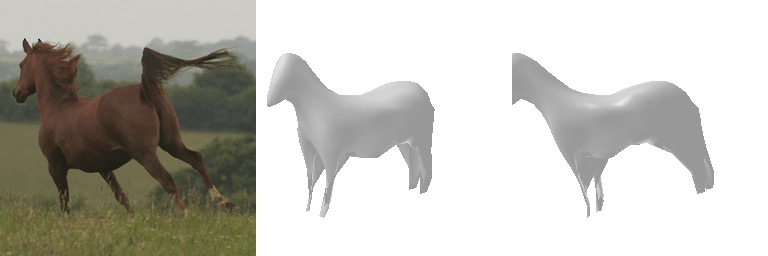} \\
\addpicsup{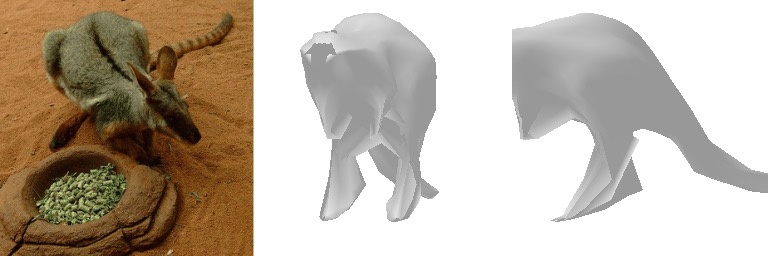} &
\addpicsup{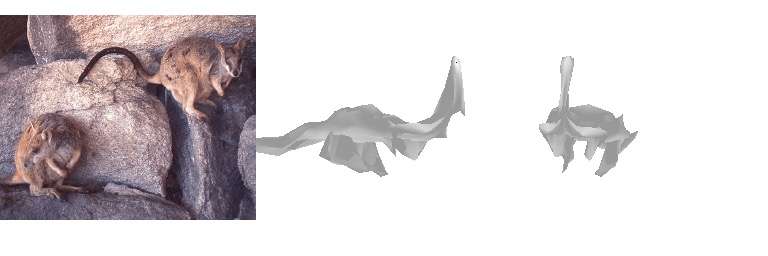} &
\addpicsup{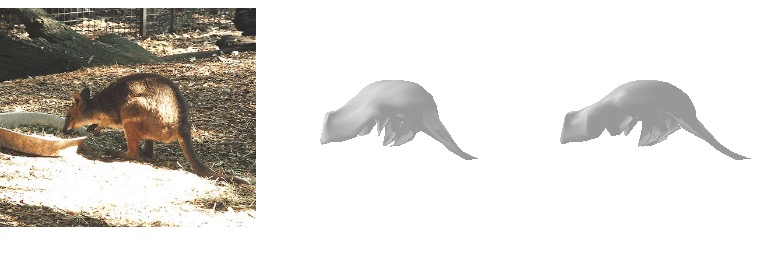} \\
\addpicsup{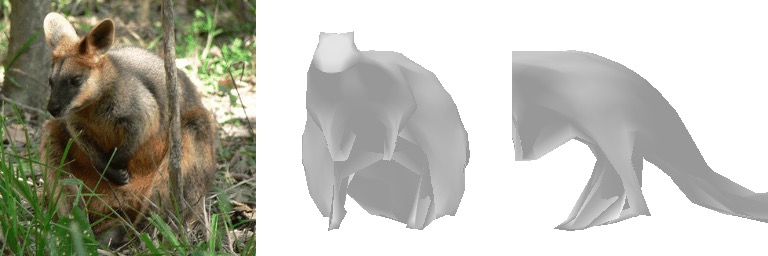} &
\addpicsup{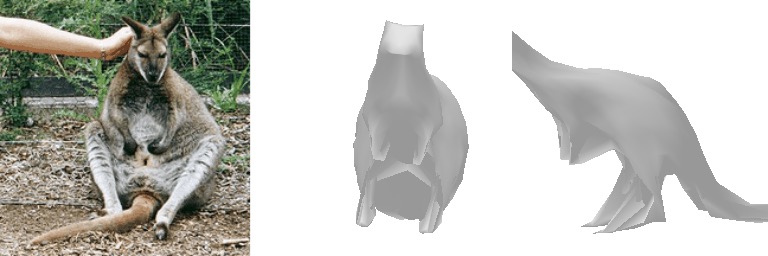} &
\addpicsup{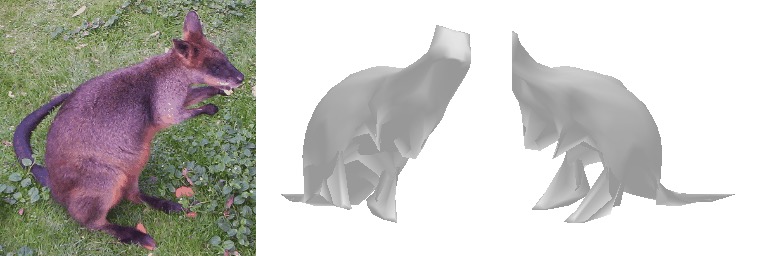} \\
\end{tabular}
}
\captionof{figure}{
\textbf{Random Results.} We show (from left to right) the input image, the inferred 3D shape from predicted view and a novel view for 6 \emph{randomly sampled} images from the test set per category.
}
\end{table*}

\begin{table*}[!t]
\setlength{\tabcolsep}{0.01em}
\renewcommand{\arraystretch}{1}
\centering
  \scalebox{0.63}{
\begin{tabular}{c@{\hskip 1em}c@{\hskip 1em}c}
\addpicsup{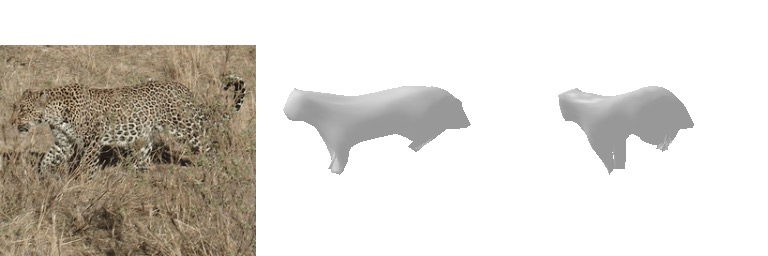} &
\addpicsup{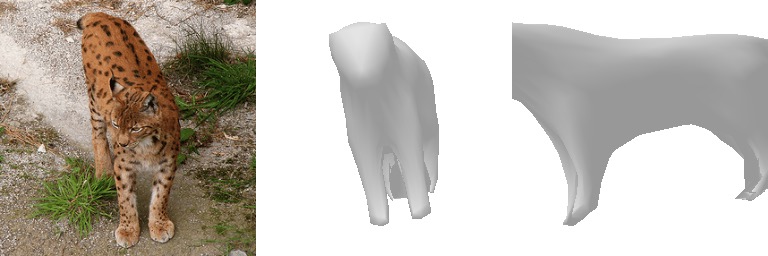} &
\addpicsup{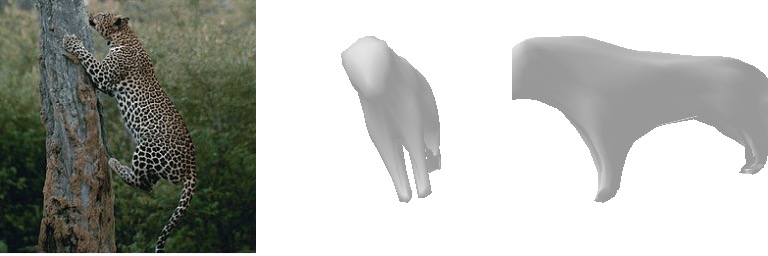} \\
\addpicsup{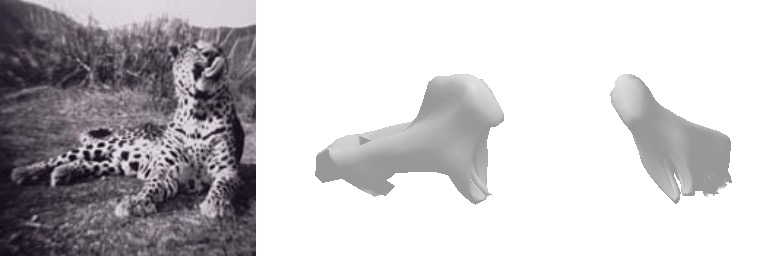} &
\addpicsup{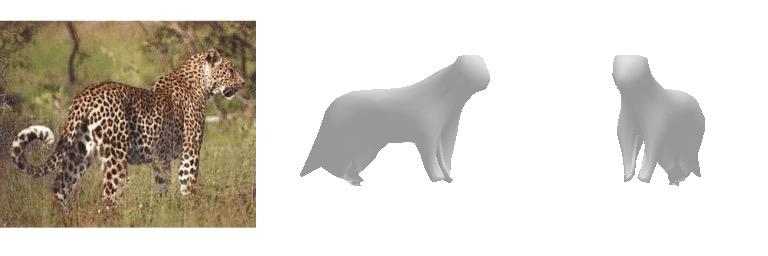} &
\addpicsup{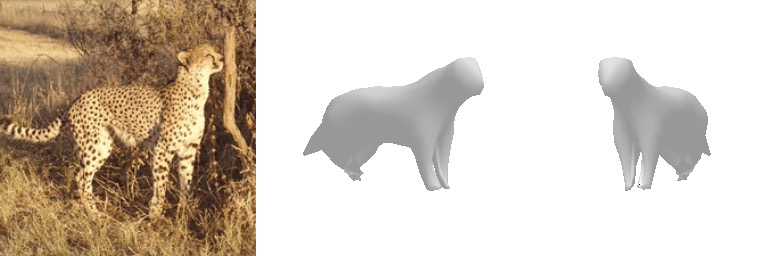} \\
\addpicsup{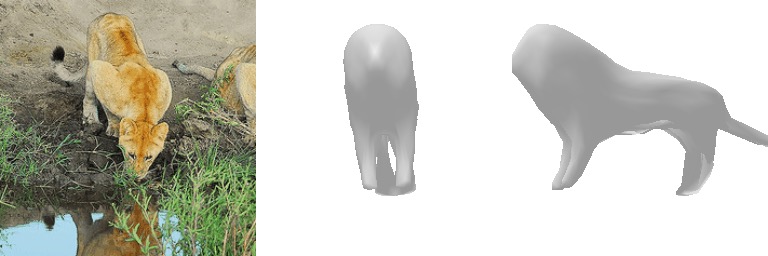} &
\addpicsup{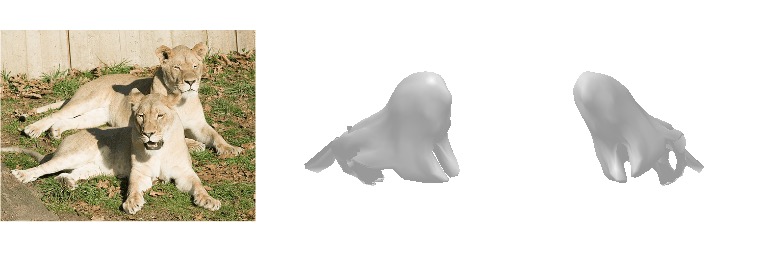} &
\addpicsup{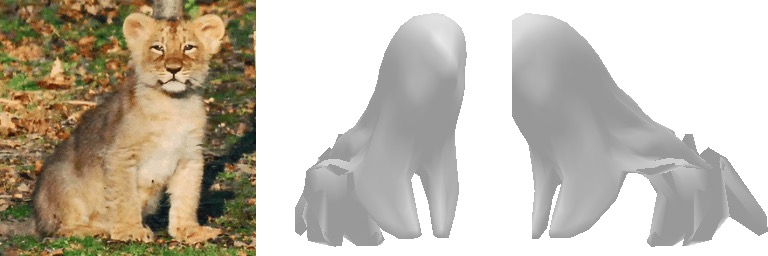} \\
\addpicsup{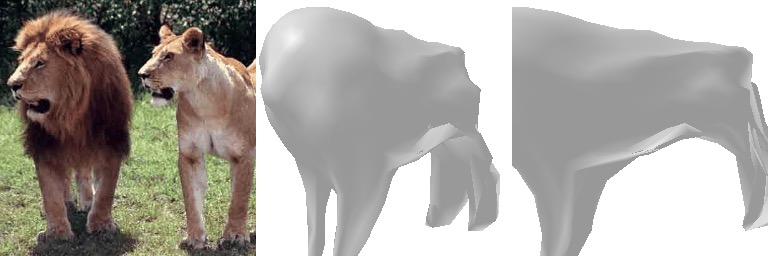} &
\addpicsup{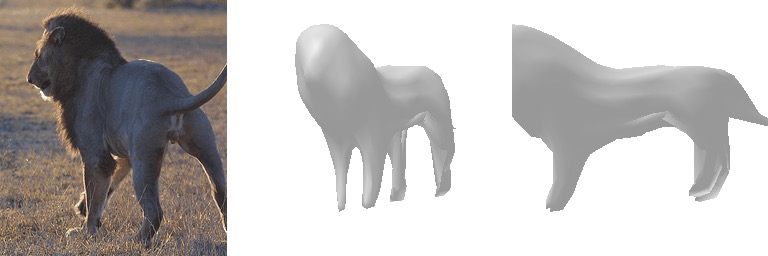} &
\addpicsup{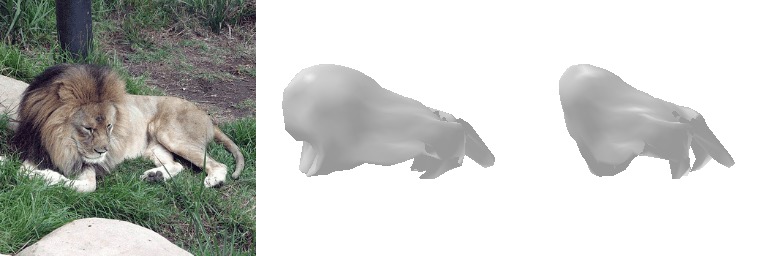} \\
\addpicsup{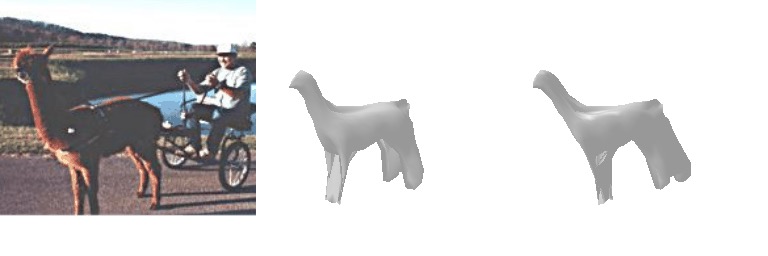} &
\addpicsup{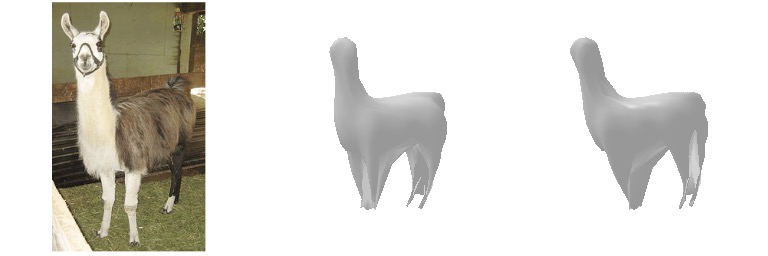} &
\addpicsup{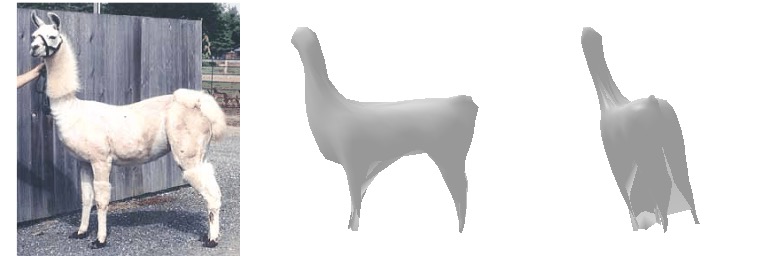} \\
\addpicsup{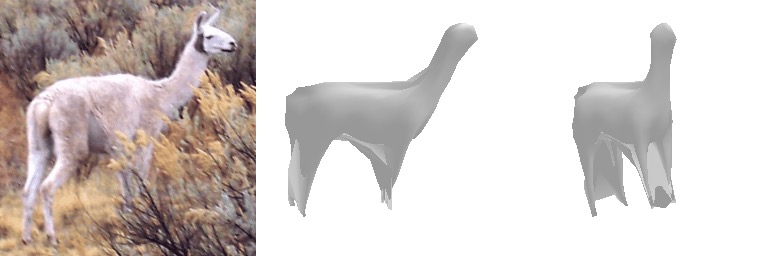} &
\addpicsup{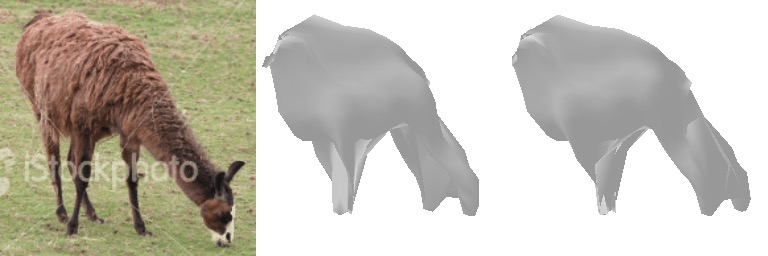} &
\addpicsup{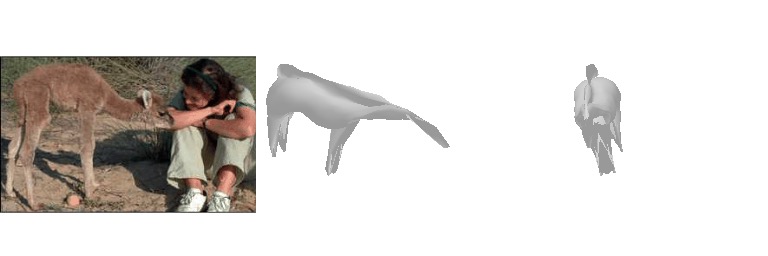} \\
\addpicsup{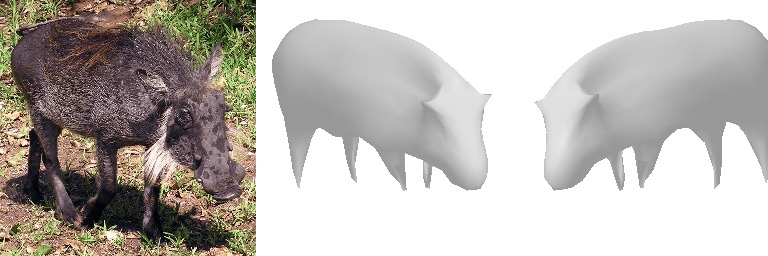} &
\addpicsup{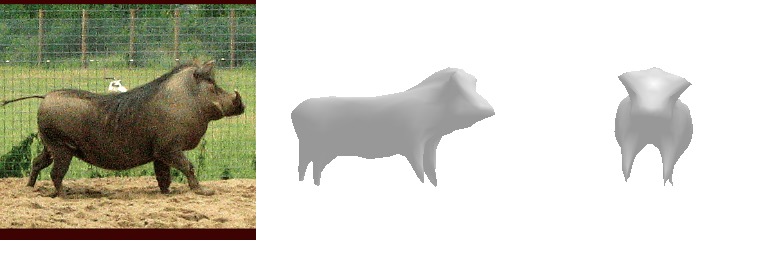} &
\addpicsup{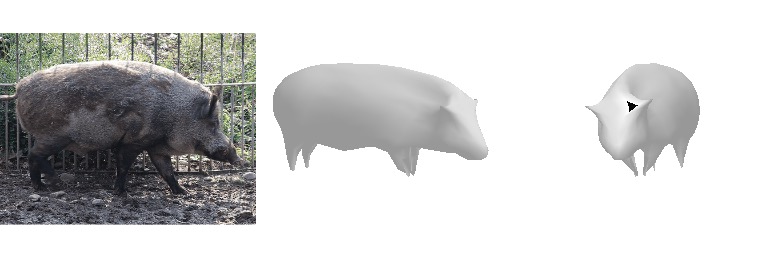} \\
\addpicsup{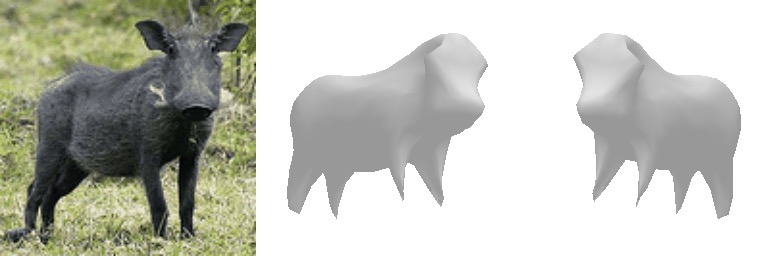} &
\addpicsup{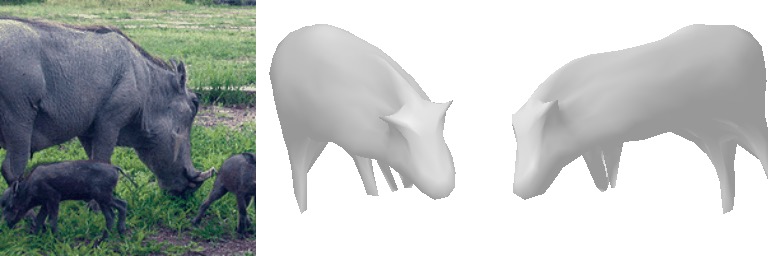} &
\addpicsup{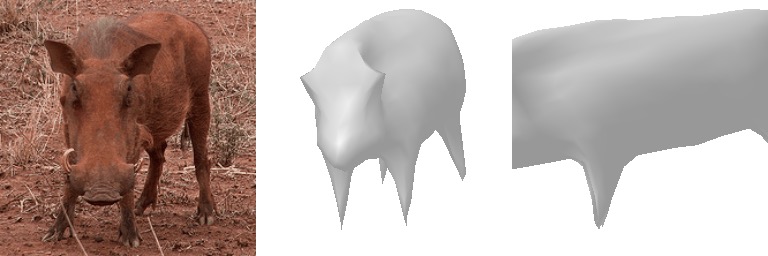} \\
\addpicsup{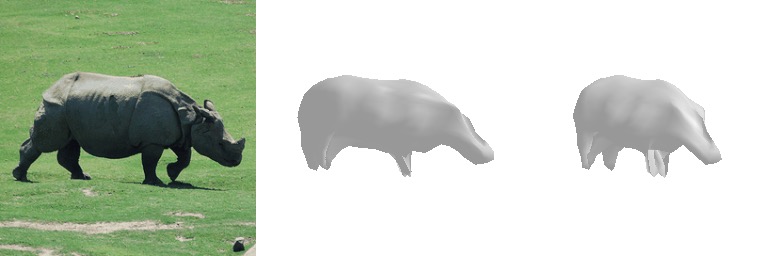} &
\addpicsup{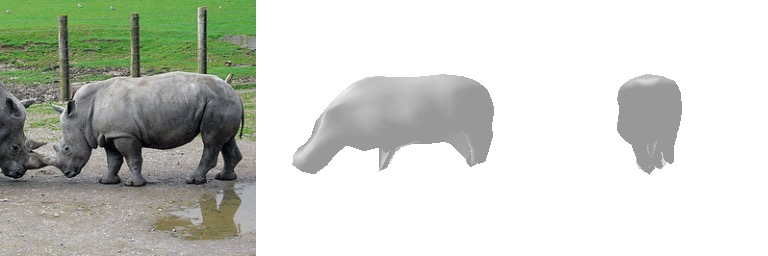} &
\addpicsup{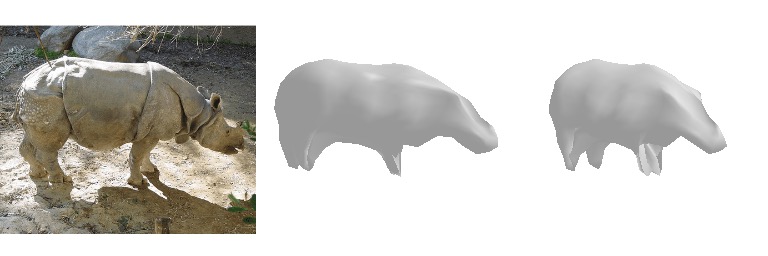} \\
\addpicsup{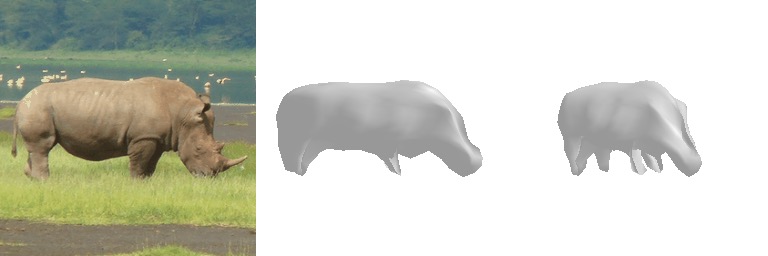} &
\addpicsup{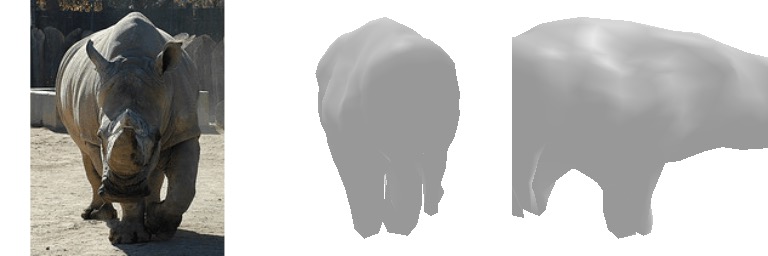} &
\addpicsup{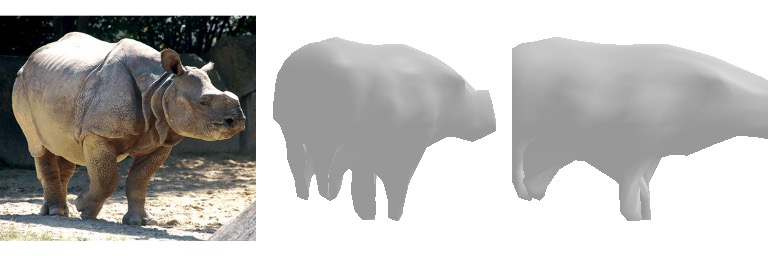} \\
\addpicsup{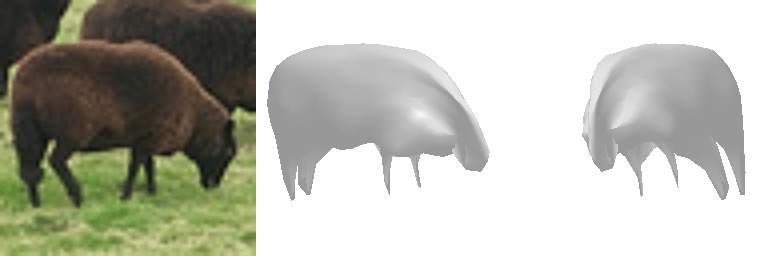} &
\addpicsup{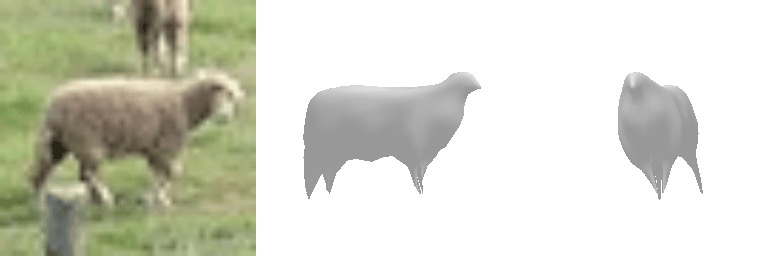} &
\addpicsup{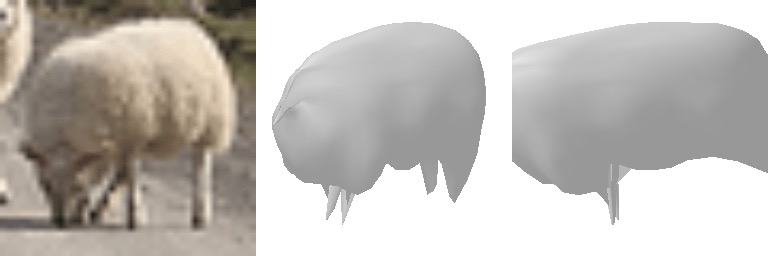} \\
\addpicsup{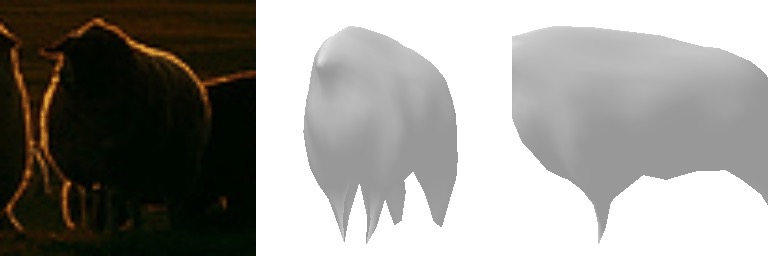} &
\addpicsup{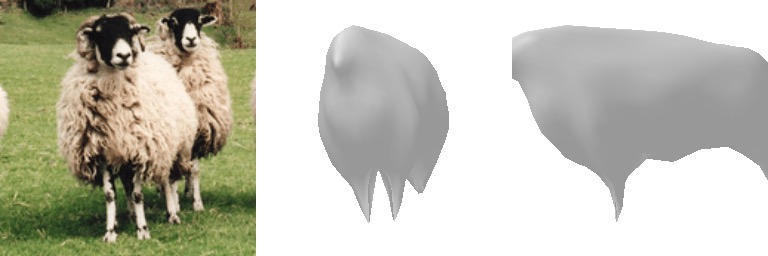} &
\addpicsup{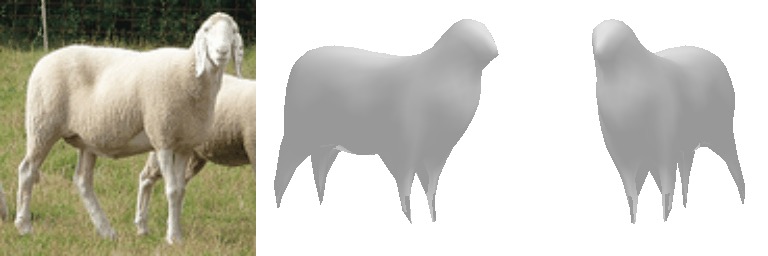} \\
\end{tabular}
}
\captionof{figure}{
\textbf{Random Results.} We show (from left to right) the input image, the inferred 3D shape from predicted view and a novel view for 6 \emph{randomly sampled} images from the test set per category.
}
\end{table*}

\begin{table*}[!t]
\setlength{\tabcolsep}{0.01em}
\renewcommand{\arraystretch}{1}
\centering
  \scalebox{0.63}{
\begin{tabular}{c@{\hskip 1em}c@{\hskip 1em}c}
\addpicsup{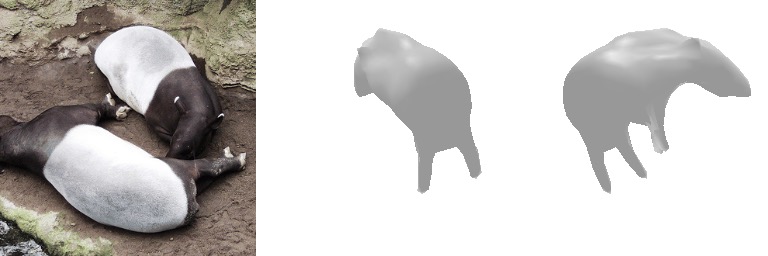} &
\addpicsup{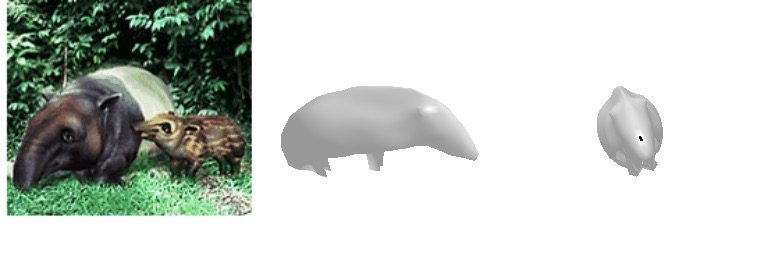} &
\addpicsup{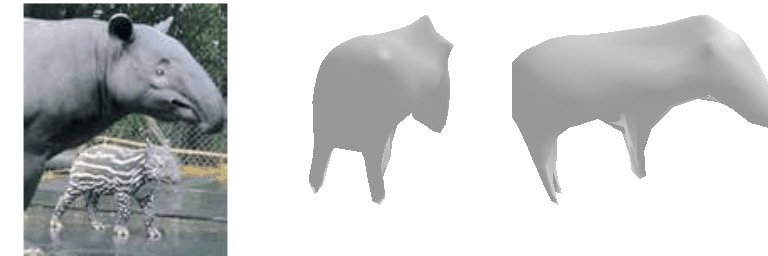} \\
\addpicsup{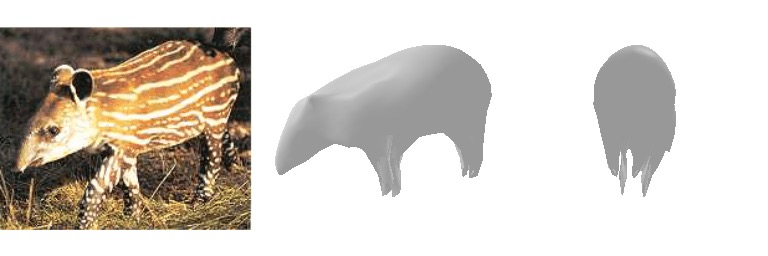} &
\addpicsup{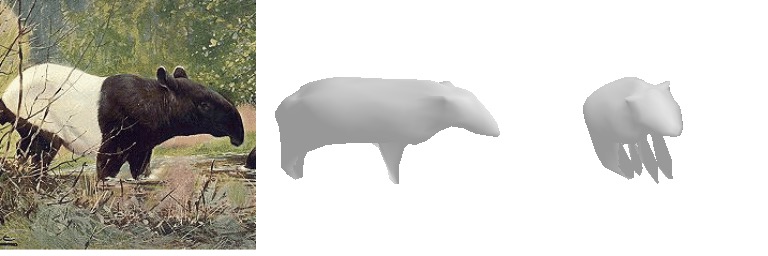} &
\addpicsup{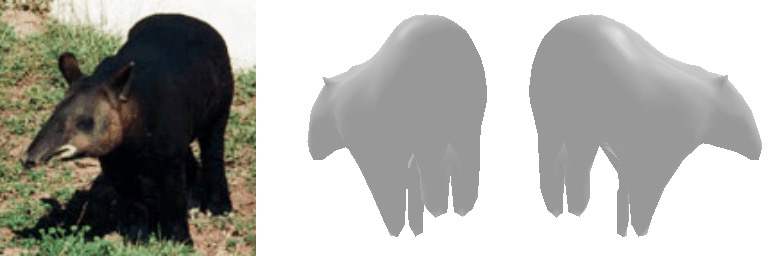} \\
\addpicsup{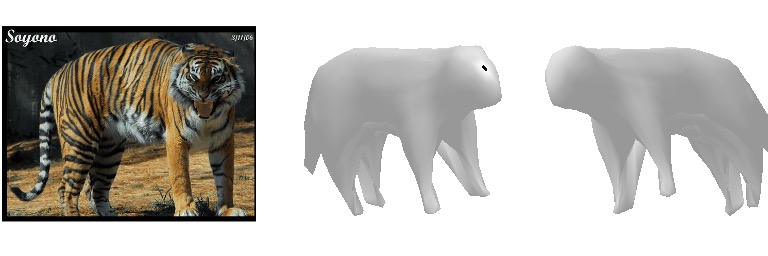} &
\addpicsup{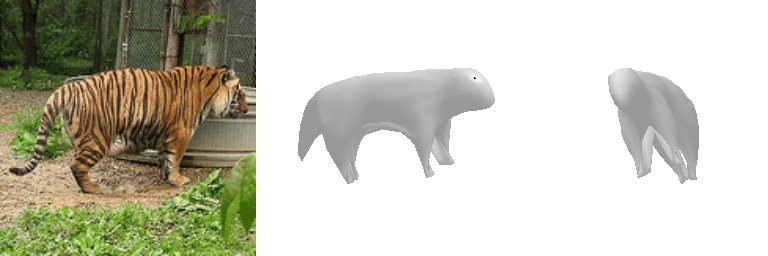} &
\addpicsup{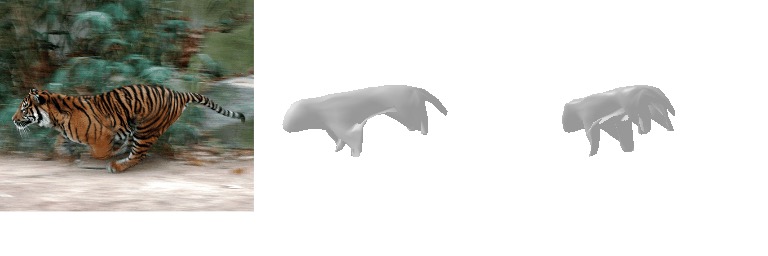} \\
\addpicsup{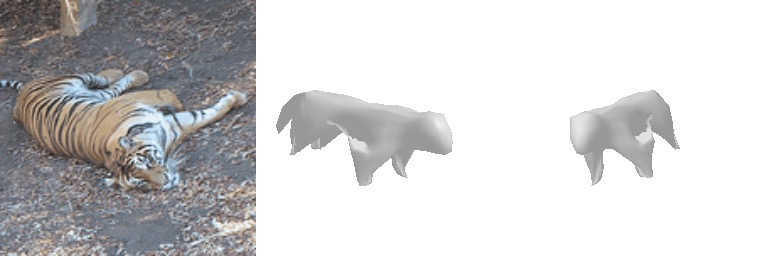} &
\addpicsup{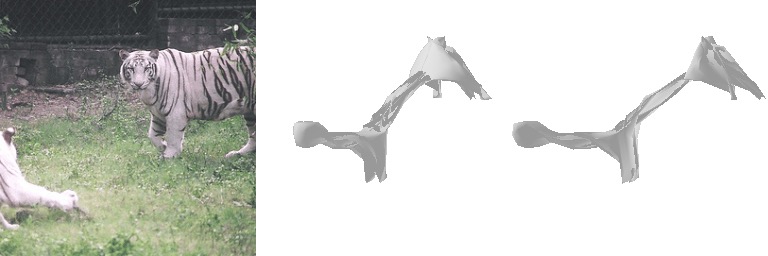} &
\addpicsup{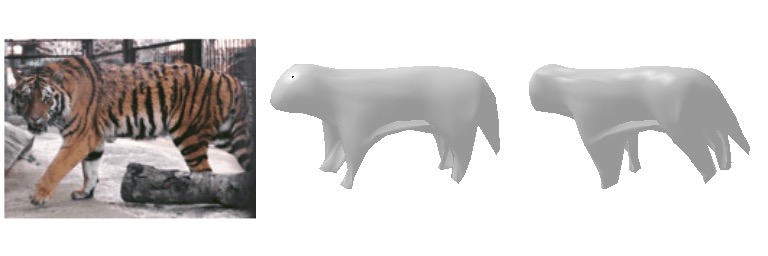} \\
\addpicsup{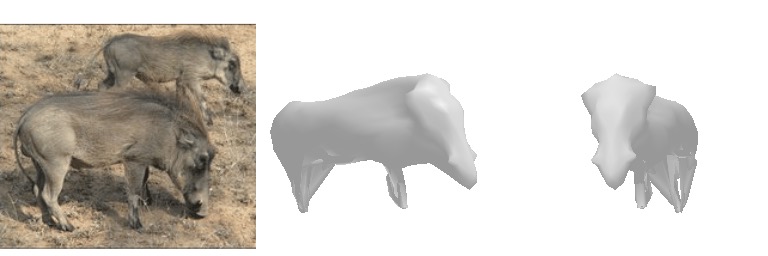} &
\addpicsup{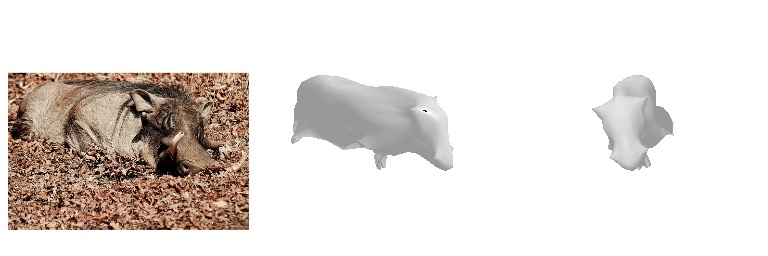} &
\addpicsup{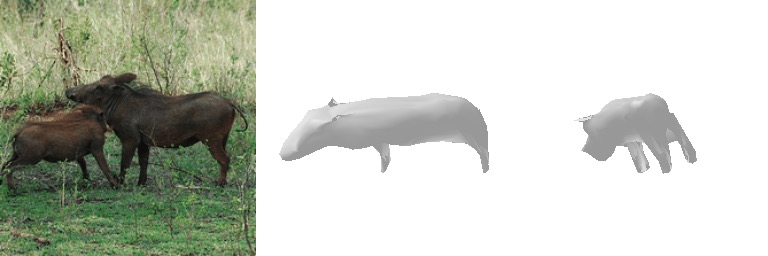} \\
\addpicsup{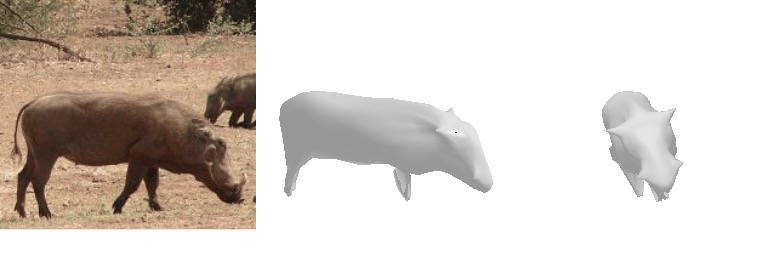} &
\addpicsup{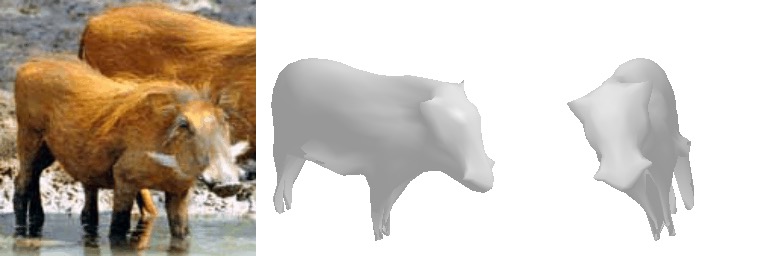} &
\addpicsup{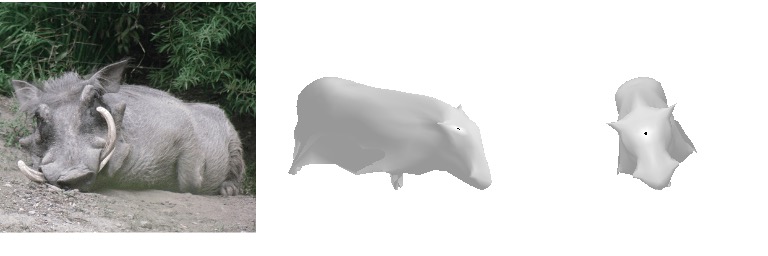} \\
\addpicsup{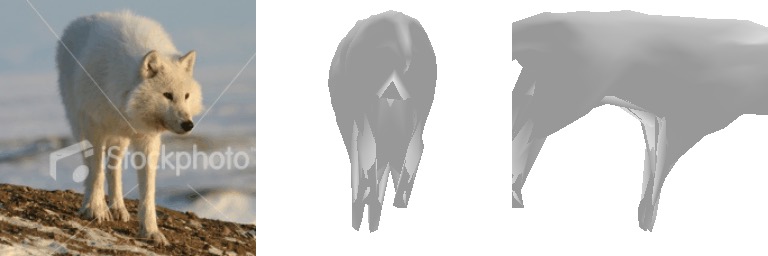} &
\addpicsup{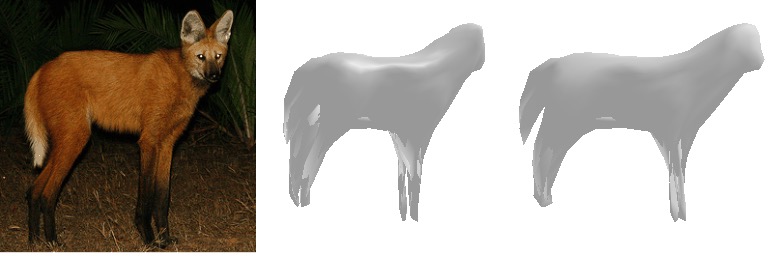} &
\addpicsup{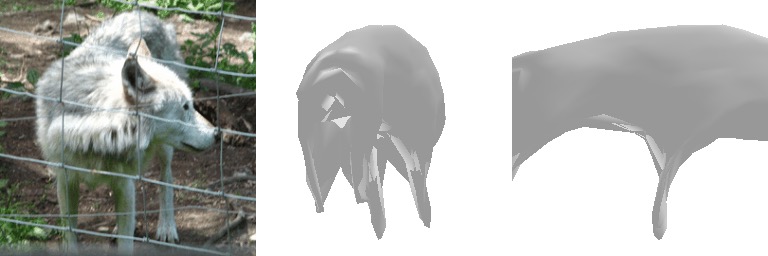} \\
\addpicsup{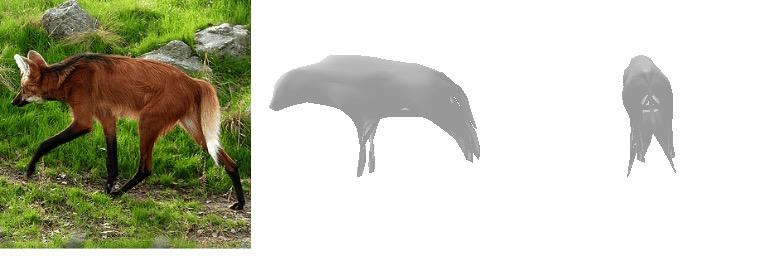} &
\addpicsup{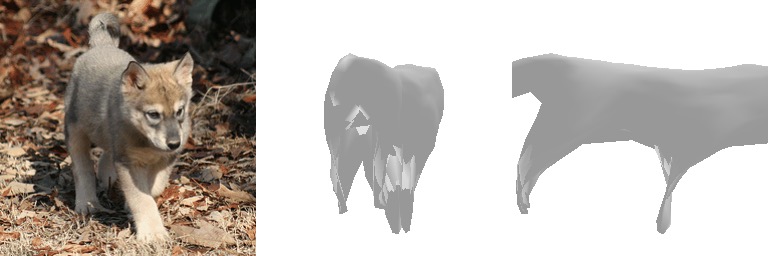} &
\addpicsup{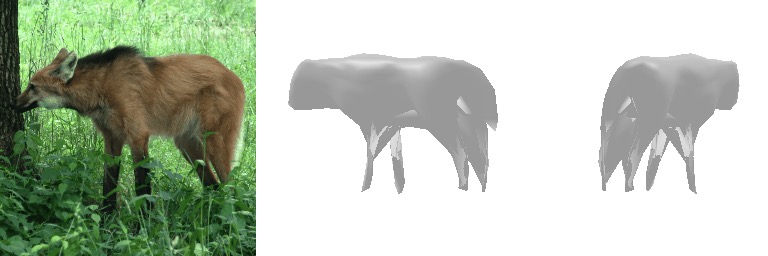} \\
\addpicsup{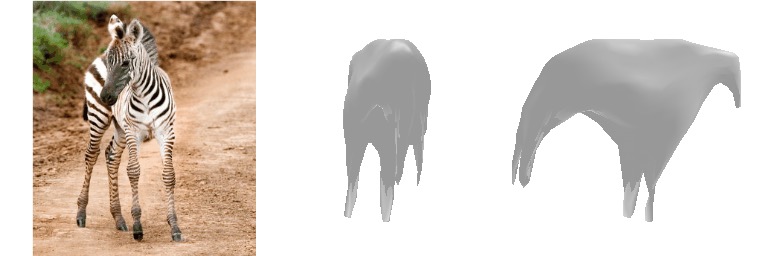} &
\addpicsup{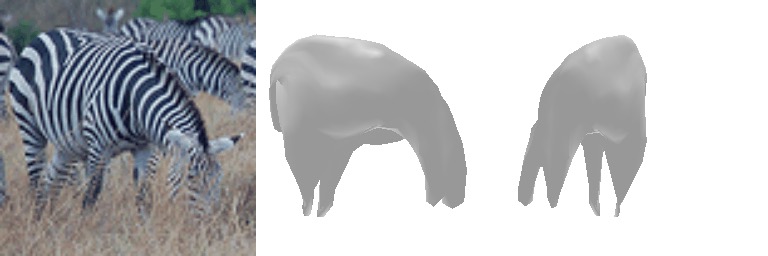} &
\addpicsup{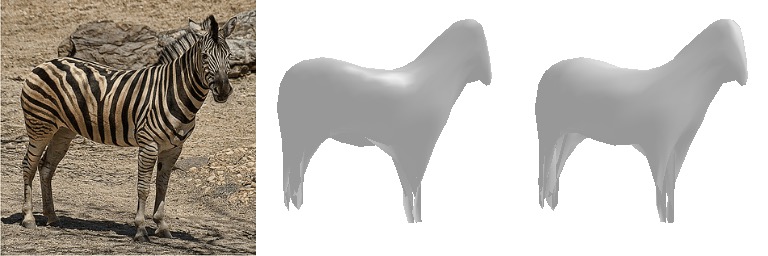} \\
\addpicsup{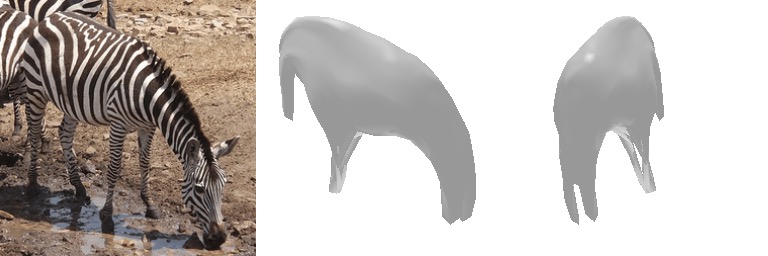} &
\addpicsup{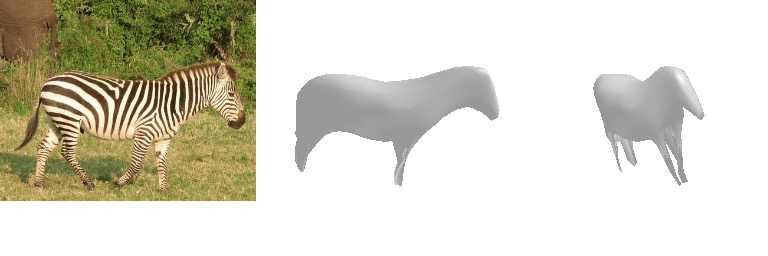} &
\addpicsup{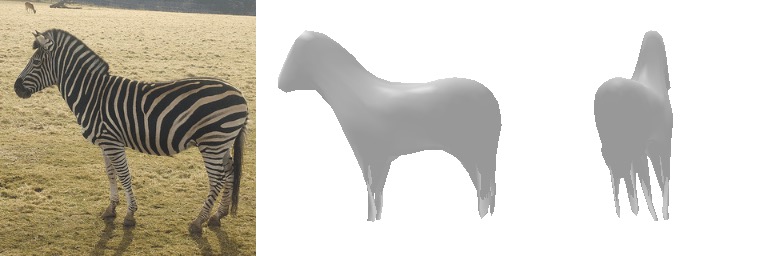} \\
\end{tabular}
}
\captionof{figure}{
\textbf{Random Results.} We show (from left to right) the input image, the inferred 3D shape from predicted view and a novel view for 6 \emph{randomly sampled} images from the test set per category.
}
\end{table*}